\documentclass[lettersize,journal]{IEEEtran}
\usepackage{amsmath,amsfonts}
\usepackage{algorithmic}
\usepackage{algorithm}
\usepackage{array}
\newcolumntype{P}[1]{>{\raggedright\arraybackslash}p{#1}}
\usepackage{textcomp}
\usepackage{stfloats}
\usepackage{url}
\usepackage{verbatim}
\usepackage{graphicx}
\usepackage{cite}
\usepackage{booktabs}
\usepackage{subfigure}
\usepackage{diagbox}
\usepackage{multirow}
\usepackage{arydshln}
\usepackage{tabularx}
\usepackage{makecell}
\begin{document}
\title{SomaAlign: Reliability-Aware Multimodal Representation Learning for Soft Robotic Grasping}

\author{Haoyuan Wang,
	Shuai Peng,
	Yongkai Liao,
	Yuxiao Sun,
	Wei Luo,
	Xinxing Chen, ~\IEEEmembership{Member,~IEEE,}
	and Jian Huang, ~\IEEEmembership{Senior Member,~IEEE}
	\thanks{Manuscript received xxxx, 2026.}
	\thanks{This work was supported in part by Hubei Provincial Technology Innovation Program under Grant 2024BAA007, and in part by National Natural Science Foundation of China under Grant 62333007 and U24A20280. \textit{(Corresponding authors: Jian Huang; Xinxing Chen.)} }
	\thanks{H. Wang, S. Peng, Y. Liao, Y. Sun, X. Chen and J. Huang are with the Hubei Key Laboratory of Brain-inspired Intelligent Systems and the Key Laboratory of Image Processing and Intelligent Control, School of Artificial Intelligence and Automation, Huazhong University of Science and	Technology (HUST), Wuhan 430074, Hubei, China. Email: {\tt\small \{why427@, pengshuai@, lyk@, sun\_yuxiao@, cxx@, huang\_jan@mail.\}hust.edu.cn}}
	\thanks{W. Luo is with Department of Innovation Center, China Ship Development and Design Center, Wuhan 430064, Hubei, China. Email: {\tt\small csddc\_weiluo@163.com}}

}

\markboth{IEEE TRANSACTIONS ON xxxx,~Vol.~xx, No.~x, xxxx~2026}%
{Shell \MakeLowercase{\textit{et al.}}: A Sample Article Using IEEEtran.cls for IEEE Journals}

\maketitle

\begin{abstract}
	Robust multimodal perception is essential for reliable soft robotic grasping, where compliant embodiment deformation, unstable contact, mechanical vibration, and heterogeneous sensor characteristics introduce modality-dependent noise and semantic ambiguity. Progress in this area is hindered by two challenges: the scarcity of soft robotic datasets with coordinated multisensory observations and the limited ability of existing alignment methods to account for varying reliability across contact-rich modalities. To address these challenges, we introduce Soma, a multimodal dataset for soft robotic grasping containing 100 objects, 1,000 continuous grasping trajectories, and 74,914 multimodal samples grouped by software acquisition cycle, spanning remote RGB, hand RGB, depth, tactile, IMU, and structured text observations. We further collect Soma-Success, an independently collected grasp-outcome dataset comprising 1,000 trials, of which 998 are retained after quality control, to assess whether frozen encoders learned from Soma can be reused for post-hoc grasp outcome classification and early grasp success prediction on a dataset unseen during representation learning. Built on Soma, we propose SomaAlign, a hierarchical multimodal representation learning framework with three complementary stages tailored to heterogeneous soft robotic signals. Masked reconstruction first derives a data-driven proxy for relative modality reliability; reliability-aware contrastive alignment then uses pair-specific adaptive temperatures to adjust alignment sharpness according to reconstruction-derived modality-pair reliability; finally, conflict-aware semantic constraints improve intra-sample cross-modal consistency while preserving global feature geometry. SomaAlign achieves an average Recall@10 of 90.66\% across 30 directed cross-modal retrieval tasks, while the learned representations further support physical attribute recognition, multimodal question answering, cross-modal generation, and cross-dataset transfer to grasp outcome prediction. On Soma-Success, frozen SomaAlign encoders with lightweight probes achieve 94.59\% post-hoc classification accuracy and 84.07\% early prediction accuracy. Together, Soma, Soma-Success, and SomaAlign provide a data and representation-learning foundation for robust and transferable multimodal perception in soft robotic manipulation.
\end{abstract}


\section{Introduction}
%

\IEEEPARstart{S}{oft} robotic grippers and hands exploit compliant structures to enable gentle interaction and manipulation of fragile or irregular objects, while their deformation-dependent contact dynamics differ fundamentally from those of rigid manipulators \cite{c4,c11,ca2}. Reliable grasping requires precise multimodal perception of contact states, material properties, and time-varying interaction dynamics; otherwise, large elastic deformation may lead to unstable holding and unreliable slip estimation \cite{c5,c6}. Multimodal sensing combines complementary signals from vision, distributed tactile arrays, and proprioceptive inertial measurements to mitigate unimodal limitations such as visual occlusion and missing contact cues, thereby improving perceptual consistency \cite{c7,c10}. Such integration is particularly important for soft robots, whose intrinsic compliance enables gentle physical interaction but also introduces deformation-dependent perceptual variations that are less pronounced in rigid platforms \cite{c12}.

Despite these advances, soft robots present two core perceptual challenges that are more pronounced than in rigid manipulators. First, structural deformation and material compliance introduce heterogeneous, modality-dependent sensor noise: tactile pressure arrays and IMUs are more susceptible to vibration during grasping, whereas RGB-D observations are generally less affected by mechanical disturbances \cite{c7}. Second, deformable contact introduces semantic ambiguity, as identical categorical grasp labels may correspond to substantially different tactile, inertial, and depth observations under different deformation magnitudes and contact configurations. Together, these factors challenge cross-modal alignment pipelines developed primarily for rigid robotic systems and may degrade conventional multimodal fusion. Robust soft-grasping perception therefore requires both coordinated multisensor data collection and representation learning explicitly adapted to modality-dependent noise and ambiguous contact semantics.

Large-scale real-world multimodal datasets provide an essential foundation for data-driven robotic manipulation, supporting imitation learning, policy generalization, and the pre-training of embodied foundation models \cite{c15,c16,c17,c18,c19,c20}. Existing tactile and visuo-tactile datasets have advanced contact-aware perception by supporting tactile localization, material classification, grasp outcome estimation, and cross-modal signal prediction \cite{ca3,ca4,ca5}. These studies demonstrate that tactile sensing captures physical interaction cues that are difficult to infer from vision alone, including contact geometry, surface properties, and object compliance. However, many existing visuo-tactile datasets are collected with rigid end-effectors or specific optical tactile sensors, where embodiment deformation and contact nonlinearity are relatively constrained. Soft grippers instead exhibit highly nonlinear kinematics, contact-state-dependent sensory responses, and heterogeneous noise distributions across visual, tactile, and inertial modalities. Consequently, datasets developed for rigid robots or specific tactile sensing systems do not fully capture the coupled multimodal dynamics intrinsic to soft grasping.

Existing multimodal tactile and grasping datasets individually support visuo-tactile learning, grasp outcome prediction, object-centric sensing, or object-in-hand pose estimation. To the best of our knowledge, no existing multimodal dataset simultaneously combines software-coordinated vision, depth, tactile, and IMU acquisition using soft grippers with multiple fingers, structured text descriptions, explicit stationary and sliding grasping states, and an independently collected grasp-outcome dataset for frozen-encoder evaluation. This combination therefore defines a specific data gap in soft robotic perception. The scarcity of such datasets stems mainly from two inherent complexities of soft systems: their high-dimensional kinematics complicate motion capture and semantic state labeling \cite{c21}, while high-fidelity simulation of soft-body contact suffers substantial Sim2Real gaps, limiting the utility of synthetic data for perceptual learning \cite{c22}. This shortage further constrains downstream perception: limited training data may reduce generalization across diverse object geometries and material properties, impairing contact-state estimation and semantic attribute recognition in unstructured environments. For soft hands in particular, grasp success depends not only on static object appearance but also on transient contact formation, spatiotemporal tactile pressure distributions, and dynamic motion stability during physical interaction---information not fully represented in existing rigid-robot datasets. Without high-quality temporally associated multimodal recordings, adapting large-scale perception models to soft manipulation remains difficult, and standardized resources for grasp outcome prediction remain limited. This yields our first core research challenge: constructing a multimodal dataset with software-coordinated sensory acquisition and structured semantic annotations that reflects the physical characteristics of soft robotic grasping, together with an independently collected dataset for evaluating the reuse of learned representations in grasp outcome prediction.

Parallel to dataset development, multimodal representation learning has advanced rapidly over the past decade. Contrastive vision--language architectures such as CLIP demonstrate that visual concepts can be embedded into transferable semantic spaces via language supervision \cite{ca6}; unified frameworks such as ImageBind further associate heterogeneous modalities, including images, text, audio, depth, thermal signals, and IMU, within a shared embedding manifold \cite{ca7}. Systematic surveys have summarized multimodal alignment and fusion paradigms across diverse modalities and general-purpose perception tasks \cite{ca12}. Despite these advances, such models are primarily optimized for generic passive perception rather than dynamic, contact-rich robotic interaction. Within the manipulation literature, recent architectures including UniTouch, AnyTouch, CLTP, and Sparsh-X integrate tactile signals with vision and language to model physical contact events \cite{c23,c24,c25,ca8}. However, these methods are built for rigid manipulators paired with optical vision-based tactile sensors, whose sensing mechanism relies on tracking localized elastomer deformation under fixed mechanical constraints. Soft robotic platforms instead employ heterogeneous sensor suites combining distributed pressure tactile arrays, inertial measurement units, and RGB-D cameras, each with distinct signal distributions and noise profiles \cite{c26,c27}.

This sensor-domain mismatch creates two critical difficulties when transferring existing alignment models to soft grippers. First, pre-trained encoders may not generalize well to the signal characteristics of flexible tactile arrays and finger-mounted IMUs. Second, conventional global alignment objectives do not explicitly account for soft-robot-specific semantic ambiguity and heterogeneous noise. Fixed or globally shared contrastive temperatures may over-constrain noisy tactile/IMU modality pairs while underutilizing reliable visual observations, leading to suboptimal embeddings and weakened preservation of physically meaningful contact information. Existing alignment frameworks also generally lack mechanisms to incorporate estimated modality reliability into pair-specific alignment strength. This forms our second core research challenge: designing a dedicated multimodal alignment framework adapted to the heterogeneous sensor characteristics and noisy, semantically ambiguous observations of soft robotic manipulation.

Collectively, these analyses reveal two coupled bottlenecks in robust multimodal perception for soft grasping: the lack of domain-specific, temporally associated multimodal data tailored to soft contact dynamics, and the absence of explicit reliability-aware mechanisms in generic cross-modal alignment pipelines for heterogeneous noisy sensory streams. Addressing either limitation alone is insufficient. We therefore present an integrated solution that combines a software-coordinated soft-grasping multimodal dataset with a hierarchical reliability-aware multimodal alignment framework. The dataset provides cycle-aligned sensory observations of contact-dependent deformation dynamics, while the alignment framework learns unified cross-modal embeddings that explicitly account for modality-dependent reliability and cross-modal semantic consistency. Together, they provide the data and algorithmic foundations for robust and transferable multimodal perception in soft robotic systems.

\begin{figure*}[tb]
	\centering
	\includegraphics[width=\textwidth]{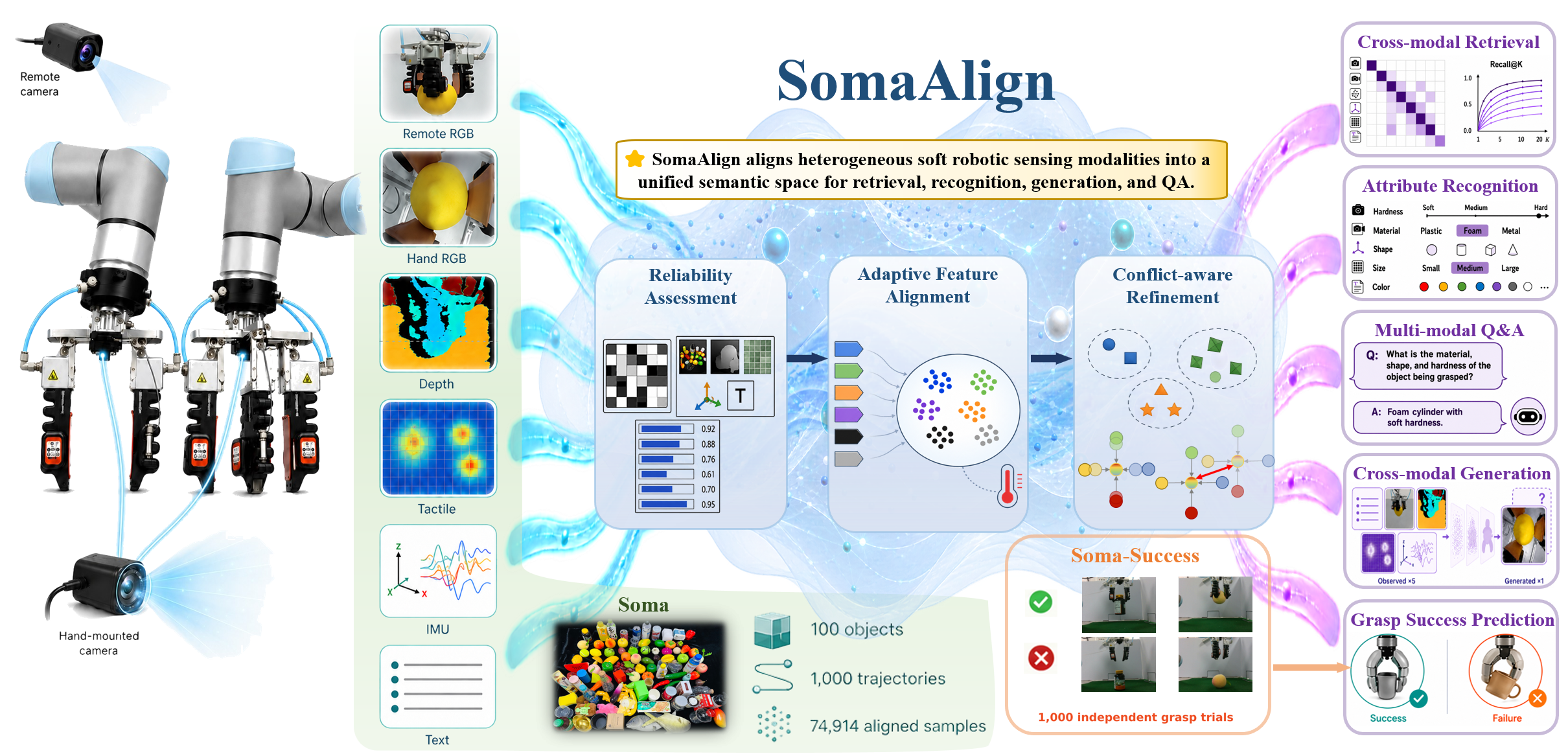} 
	\caption{Overview of Soma, Soma-Success, and the SomaAlign framework. The soft robotic platform acquires six coordinated modalities: Remote RGB, Hand RGB, Depth, Tactile, IMU, and Text. Soma comprises 100 objects, 1,000 trajectories, and 74,914 aligned multimodal samples, while Soma-Success contains 1,000 collected grasp	trials, of which 998 valid trials are retained for downstream evaluation. SomaAlign learns a unified semantic representation through Reliability Assessment, Adaptive Feature Alignment, and Conflict-aware Refinement. The learned representation supports cross-modal retrieval, physical attribute recognition, multimodal question answering, cross-modal generation, and frozen-encoder grasp success prediction.}
	\label{fig:teaser}
\end{figure*}

Our main contributions are as follows:
\begin{enumerate}
	\item We introduce Soma (SOft-robotic Multimodal Acquisition), which, to the best of our knowledge, is the first six-modality dataset dedicated to unified representation learning for	soft robotic grasping. Soma combines cycle-aligned remote RGB, hand RGB, depth, distributed tactile pressure, and IMU observations with structured text annotations, covering stationary and sliding grasp states.
	We further introduce Soma-Success, an independently collected grasp-outcome dataset excluded from representation learning, enabling frozen-encoder evaluation of post-hoc outcome classification and early success prediction.
	
	\item We propose SomaAlign, which, to the best of our knowledge, is the first multimodal alignment framework to jointly incorporate modality reliability and cross-modal conflict into unified representation learning for soft robotic grasping. Reconstruction-derived reliability regulates modality-pair-specific contrastive temperatures. Subsequent refinement uses batch-level conflict modulation to adjust intra-sample contraction and frozen Stage-II reference centers to limit excessive geometric drift. The framework connects these mechanisms through three progressive training stages and retains only the modality encoders at inference.
	
	\item We establish a systematic evaluation protocol for multimodal representations in soft robotic grasping, covering cross-modal correspondence, physical-attribute recognition, and frozen-encoder reuse on independently collected grasp-outcome data. Baseline comparisons, sensor-corruption tests, and component ablations assess the proposed alignment mechanisms, while post-hoc classification and early prediction examine their downstream utility with complete and partial observations.
\end{enumerate}

The implementation code of SomaAlign is publicly available at https://github.com/Hijackoo/SomaAlign to facilitate reproducibility. The Supplementary Information provides additional experimental details and analyses, while accompanying videos illustrate the proposed approach and representative experimental results. An overview of Soma, Soma-Success, and the SomaAlign framework is presented in Fig. \ref{fig:teaser}. The remainder of this paper is organized as follows: Section II introduces the Soma and Soma-Success datasets; Section III presents the SomaAlign methodology; Section IV reports the experimental results and discussion; and Section V concludes the paper.

\section{Soma and Soma-Success Datasets}

Soma is a software-coordinated multimodal dataset for soft robotic grasping, comprising 74,914 cycle-aligned samples collected from 100 objects and 1,000 grasping trajectories using two- and three-finger pneumatic soft grippers. Each sample contains remote RGB, hand-mounted RGB, depth, tactile pressure, IMU, and structured text observations with stationary or sliding grasp-state annotations. We additionally collect Soma-Success as an independent grasp-outcome evaluation set; it is excluded from SomaAlign representation learning and used only to assess the downstream reusability of frozen SomaAlign encoders.

\subsection{Comparison with Existing Multimodal Tactile and Grasping Datasets}

We compare Soma with representative multimodal tactile and grasping datasets, including SoftGrasp \cite{ca16}, The Feeling of Success \cite{c19}, Touch and Go \cite{c20}, VisGel \cite{ca4}, TVL \cite{c15}, the ObjectFolder series \cite{c16,c17,c18}, and VinT-6D \cite{ca17}. Table~\ref{tab:dataset_comparison} summarizes their characteristics along six dimensions relevant to soft robotic perception.

\begin{table*}[tb]
	\centering
	\caption{Comparison of representative multimodal tactile and grasping datasets along six dimensions relevant to soft robotic perception.}
	\label{tab:dataset_comparison}
	\small
	\renewcommand{\arraystretch}{1.18}
	\setlength{\tabcolsep}{2.8pt}
	
	\begin{tabular}{
			>{\centering\arraybackslash}p{1.70cm}
			>{\centering\arraybackslash}p{2.35cm}
			>{\centering\arraybackslash}p{2.65cm}
			>{\centering\arraybackslash}p{2.15cm}
			>{\centering\arraybackslash}p{2.15cm}
			>{\centering\arraybackslash}p{2.30cm}
			>{\centering\arraybackslash}p{2.45cm}}
		\toprule
		
		\textbf{Dataset} &
		\shortstack{\textbf{Acquisition}\\\textbf{Coordination}} &
		\textbf{Modalities} &
		\textbf{Embodiment} &
		\shortstack{\textbf{Sliding}\\\textbf{States}} &
		\shortstack{\textbf{Text}\\\textbf{Annotation}} &
		\shortstack{\textbf{Independent}\\\textbf{Evaluation Set}} \\
		
		\midrule
		
		SoftGrasp \cite{ca16} &
		Not reported &
		RGB, joint angles, and joint torques &
		Articulated dexterous hand &
		Not explicitly included &
		None &
		No \\
		
		\addlinespace[2pt]
		
		The Feeling of Success \cite{c19} &
		Not reported &
		RGB and dual GelSight tactile images &
		Parallel-jaw gripper &
		Not explicitly included &
		None &
		No \\
		
		\addlinespace[2pt]
		
		Touch and Go \cite{c20} &
		Timestamped visual and tactile streams &
		Egocentric RGB and GelSight tactile video &
		Human-held tactile sensor &
		No grasp-state annotation &
		Material labels only &
		No \\
		
		\addlinespace[2pt]
		
		VisGel \cite{ca4} &
		Paired visual--tactile sequences &
		RGB and GelSight tactile images &
		Robot-mounted tactile sensor &
		No explicit grasp-state annotation &
		None &
		No \\
		
		\addlinespace[2pt]
		
		TVL \cite{c15} &
		Paired vision--touch samples &
		RGB, tactile images, and language &
		Human-held tactile sensor &
		No grasp-state annotation &
		Natural-language descriptions &
		No \\
		
		\addlinespace[2pt]
		
		ObjectFolder series \cite{c16,c17,c18} &
		Object-level aligned data &
		Vision, touch, and audio &
		Object-centric measurement and simulation &
		No explicit grasp-state annotation &
		Object metadata only &
		No separate subset \\
		
		\addlinespace[2pt]
		
		VinT-6D \cite{ca17} &
		Calibrated and aligned samples &
		RGB, depth, touch,\newline
		proprioception, and 6D pose &
		Multi-finger robotic hands &
		No explicit sliding-state annotation &
		None &
		No separate subset \\
		
		\midrule
		
		\textbf{Soma / Soma-Success} &
		\textbf{Sequential software polling} &
		\textbf{Remote RGB, hand RGB, depth,}\newline
		\textbf{tactile pressure, IMU, and text} &
		\textbf{Two- and three-finger}\newline
		\textbf{pneumatic soft grippers} &
		\textbf{Stationary and}\newline
		\textbf{sliding states} &
		\textbf{Structured attribute and}\newline
		\textbf{state descriptions} &
		\textbf{Independent Soma-Success}\newline
		\textbf{evaluation set} \\
		
		\bottomrule
	\end{tabular}
\end{table*}

Table~\ref{tab:dataset_comparison} highlights the distinct characteristics of Soma for soft robotic perception. Soma combines software-coordinated multimodal observations from two pneumatic soft-gripper configurations with explicit stationary and sliding grasp states and structured semantic descriptions. Soma-Success further provides an independently collected grasp-outcome dataset for evaluating the downstream reusability of Soma-trained encoders. Together, these characteristics make the proposed datasets well suited for studying multimodal perception under the deformation, contact variability, and heterogeneous sensing conditions of soft robotic grasping.

\subsection{Data Collection}

\textbf{Acquisition System:}
To establish cycle-level correspondence among heterogeneous sensory observations, we develop a software-coordinated acquisition system centered on a pneumatic soft robotic hand. All modalities are queried sequentially within a unified host-side polling loop, and observations returned in each cycle are grouped into one multimodal sample, providing deterministic cycle-level association across modalities.

\begin{figure*}[tb]
	\centering
	\includegraphics[width=\textwidth]{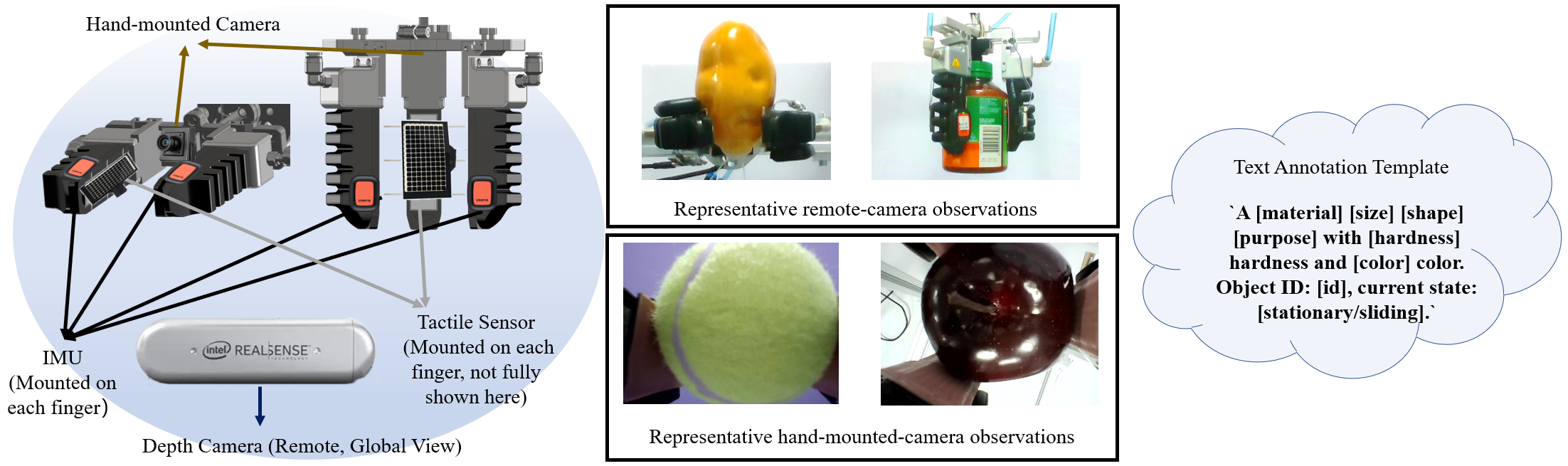}
	\caption{Overview of the Soma acquisition system. A unified software polling loop coordinates remote RGB-D sensing, hand-mounted RGB cameras, finger-mounted IMUs, and flexible tactile pressure arrays. Structured text descriptions are generated from experimental metadata using predefined templates. Representative sensor outputs and an example text annotation are shown on the right.}
	\label{DataHard}
\end{figure*}

\textbf{Sensing Configuration:}
The perception unit comprises four sensor groups. An Intel RealSense D435i depth camera ($640\times480$) captures remote RGB images and depth maps of the global scene. A hand-mounted industrial camera (JIERUIWEITONG GF100, 2.5~mm distortion-free lens, $320\times240$, $100^{\circ}$ FOV) captures local object and fingertip views. One Xsens DOT sensor is mounted on each soft finger to record three-axis Euler angles, acceleration, and angular velocity, together with a timestamp. Only the nine sensor channels are used as encoder inputs. HiT Palm flexible pressure arrays are mounted on the finger contact surfaces, with each array producing an $8\times14$ tactile pressure map. The complete acquisition system and representative modality outputs are shown in Fig.~\ref{DataHard}.

\textbf{Execution System:}
The execution unit employs pneumatic soft robotic hands actuated by regulated air pressure. An oil-free air compressor (Denair DW35, China) supplies air to an iPCU2-HMN driver (ROUCHU, China), which regulates the output pressure to control the FM-A6V5/LS1[P] soft actuators (ROUCHU, China) \cite{c7}. Two gripper configurations are used: a two-finger opposing configuration (2F) and a three-finger enveloping configuration (3F).

\begin{figure}[tbp]
	\centering
	\includegraphics[width=\linewidth]{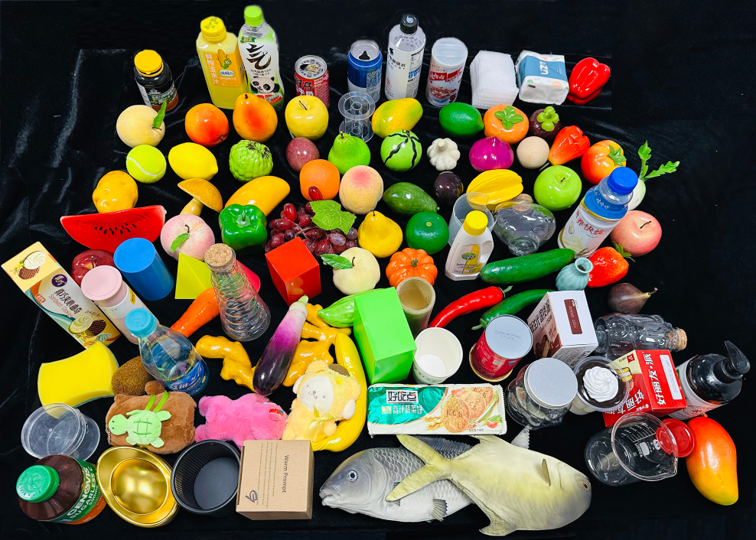}
	\caption{Object collection in Soma. The 100 everyday objects cover diverse materials, shapes, sizes, hardness levels, colors, and functional purposes.}
	\label{fig:dataset_overview}
\end{figure}

\textbf{Object and Trajectory Collection:}
We select 100 everyday objects spanning diverse materials, shapes, sizes, hardness levels, colors, and functional purposes, as shown in Fig.~\ref{fig:dataset_overview}. The collection includes plastic, glass, metal, ceramic, wood, fabric, foam, and sponge objects with both regular and irregular geometries.

To capture the transition from stationary grasping to sliding and unstable release, we design a controllable sliding device. The lower end of each object is connected by a fishing line to a motorized rod through a fixed pulley. During acquisition, the rod pulls the grasped object downward until it approaches release or fully drops from the soft fingers. Each object is tested in 10 independent trajectories, with five using the 2F gripper and five using the 3F gripper, yielding 1,000 continuous grasping trajectories in total.

\subsection{Annotation and Preprocessing}

\textbf{Temporal Segmentation:}
Each continuous trajectory is segmented using a non-overlapping temporal window with both the window size and step size set to three acquisition cycles. This three-cycle window provides compact local temporal context, retaining short-term variations across consecutive multimodal observations while reducing redundancy from densely sampled neighboring cycles. After segmentation, Soma contains 74,914 cycle-aligned multimodal samples, each comprising remote RGB, hand-mounted RGB, depth, tactile pressure, IMU, and structured text fields.

\textbf{Semantic Annotation:}
Structured text descriptions are deterministically generated from experimental metadata to ensure semantic accuracy and consistency. For each sample, material, size, shape, purpose, hardness, color, object identity, and grasping state are mapped to standardized keywords, with grasping state represented by two labels: stationary and sliding. The final description follows the template: ``A [material] [size] [shape] [purpose] with [hardness] hardness and [color] color. Object ID: [id], current state: [stationary/sliding].'' This protocol yields sample-level descriptions for individual three-cycle windows and experiment-level summaries for complete trajectories.

\textbf{Dataset Split:}
Soma is divided into training, validation, and test sets at a ratio of 60\%, 20\%, and 20\%, respectively. The split is performed by experiment ID rather than by individual sample, ensuring that all temporal windows from the same object-pulling trajectory remain in the same subset and preventing temporally correlated windows from appearing across training and evaluation sets.

\subsection{Dataset Statistics and Quality Control}

To quantify the annotation diversity of Soma, we analyze the grasping state and six semantic attributes using Shannon entropy and normalized evenness. For an annotation dimension with $K$ categories, let $p_k$ denote the proportion of samples in the $k$-th category. We compute the Shannon entropy $H$, its theoretical maximum $H_{\max}$, and the normalized evenness $J$ as

\begin{equation}
	\begin{aligned}
		H &= -\sum_{k=1}^{K} p_k \ln p_k, 
		H_{\max} &= \ln K, 
		J &= \frac{H}{H_{\max}}.
	\end{aligned}
\end{equation}
Here, $0\leq J\leq1$, with larger values indicating a more even category distribution. The six semantic attributes contain 75 attribute values in total. As shown in Table~\ref{tab:soma_annotation_distribution}, shape, hardness, and color exhibit relatively high evenness, whereas material and purpose are more concentrated in common everyday categories; size remains moderately balanced. The grasping-state distribution is moderately skewed toward sliding ($J=0.816$), while both stationary and sliding states remain substantially represented. This imbalance reflects the collection process, in which the stationary phase is generally shorter than the subsequent sliding phase. We therefore retain the original binary state distribution rather than artificially resampling the data or introducing unannotated intermediate states.

\begin{table}[tb]
	\centering
	\caption{Diversity and distribution evenness of the grasping-state and semantic annotations in Soma. $K$ denotes the number of categories, $H$ is the Shannon entropy, $H_{\max}=\ln K$, and $J=H/H_{\max}$ is the normalized evenness.}
	\label{tab:soma_annotation_distribution}
	\begin{tabular}{ccccc}
		\toprule
		\textbf{Annotation Dimension} &
		$\boldsymbol{K}$ &
		$\boldsymbol{H}$ &
		$\boldsymbol{H_{\max}}$ &
		$\boldsymbol{J}$ \\
		\midrule
		Material & 10 & 1.549 & 2.303 & 0.673 \\
		Hardness & 5 & 1.419 & 1.609 & 0.882 \\
		Size & 6 & 1.330 & 1.792 & 0.742 \\
		Shape & 14 & 2.184 & 2.639 & 0.828 \\
		Color & 28 & 2.929 & 3.332 & 0.879 \\
		State & 2 & 0.566 & 0.693 & 0.816 \\
		Purpose & 12 & 1.667 & 2.485 & 0.671 \\
		\bottomrule
	\end{tabular}
\end{table}

%


The 2F and 3F subsets contain 37,492 (50.05\%) and 37,422 (49.95\%) samples, respectively, and both cover all 100 objects. Before dataset splitting, all segmented samples are checked for modality completeness, file readability, expected dimensions, and numerical validity; only samples containing all six valid modality fields are retained. Detailed gripper-wise distribution and object-level evenness statistics are provided in the Supplementary Information.

\subsection{Soma-Success: Independent Grasp Outcome Evaluation Set}
\label{sec:soma_success}

\textbf{Collection Protocol:}
Soma-Success is independently collected to evaluate whether Soma-trained encoders can be reused for grasp outcome prediction on data unseen during representation learning. It is excluded from SomaAlign training and used only for frozen-encoder evaluation with lightweight downstream probes. Using a UR-mounted soft robotic gripper, we collect 1,000 trials across 100 everyday objects, with each object tested five times under the 2F configuration and five times under the 3F configuration, yielding 500 trials per configuration.

\begin{figure*}[tb]
	\centering
	\includegraphics[width=\textwidth]{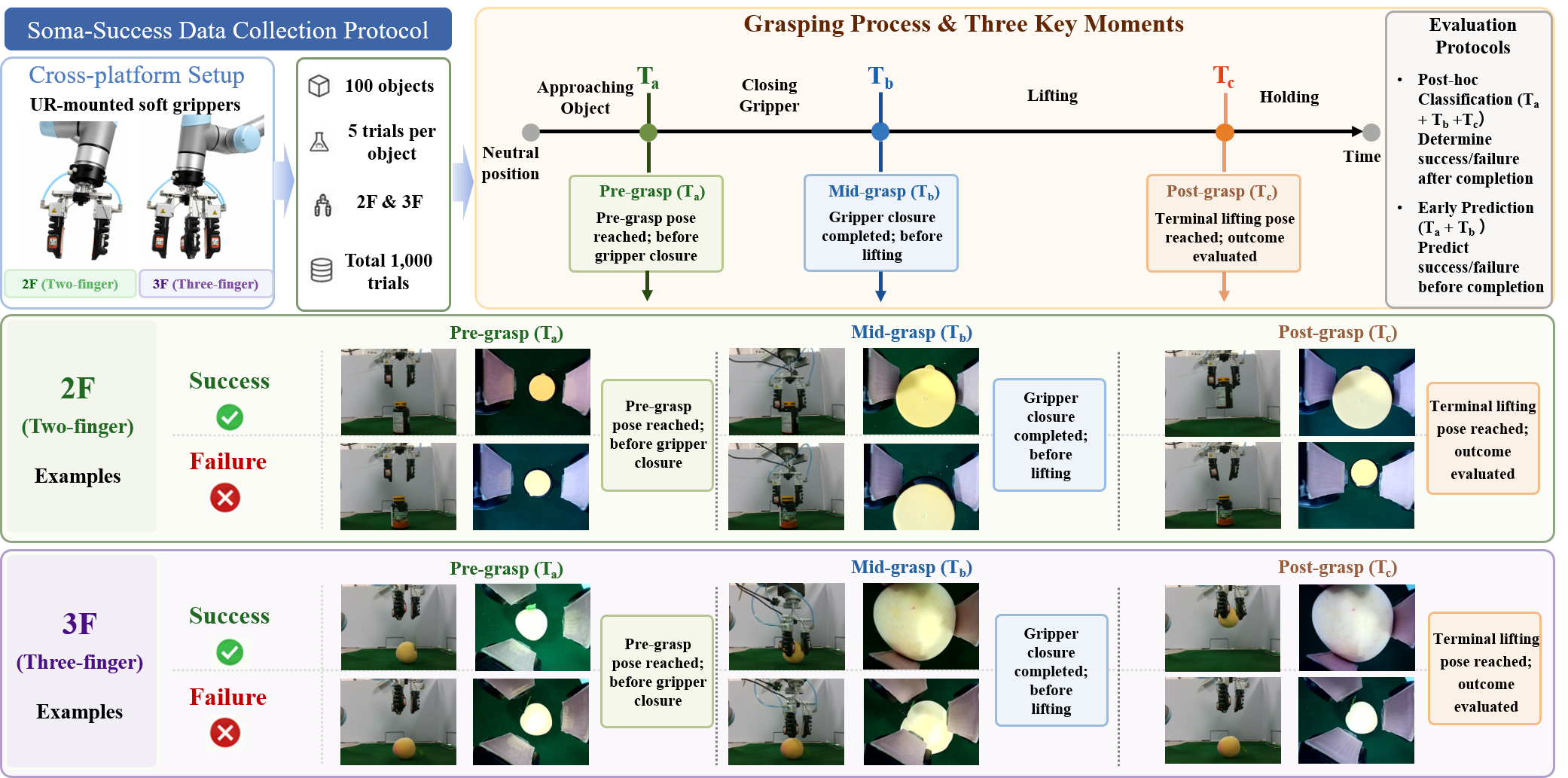}
	\caption{Soma-Success collection protocol. The action-triggered snapshots $T_a$, $T_b$, and $T_c$ correspond to the pre-grasp, closed-gripper, and terminal-lifting stages, respectively. The same snapshot-selection rule is applied to successful and failed trials. Representative examples are shown for the two-finger (2F) and three-finger (3F) configurations.}
	\label{fig:soma_success_protocol}
\end{figure*}

\textbf{Key-Snapshot Protocol:}
Each trial contains three action-triggered snapshots rather than uniformly sampled frames. $T_a$ is the last complete acquisition cycle after the manipulator reaches the pre-grasp pose and before the gripper-closing command. $T_b$ is the first complete cycle after the gripper reaches its commanded closing state and establishes object contact, but before lifting begins. $T_c$ is the first complete cycle after the manipulator reaches the predefined terminal lifting pose. Because the snapshots are action-triggered, their temporal intervals may vary across trials. If slip or release occurs during lifting, $T_c$ remains defined by the terminal lifting pose rather than the slip onset, preventing outcome-dependent snapshot selection. This protocol supports post-hoc classification using $(T_a,T_b,T_c)$ and early prediction using only $(T_a,T_b)$. The complete procedure, snapshot definitions, and representative successful and failed trials are shown in Fig.~\ref{fig:soma_success_protocol}.

\textbf{Trial Representation and Outcome Labels:}
At each key snapshot, we record frontal RGB, top-view RGB, depth, tactile pressure, IMU, and structured text observations. Each piezoresistive tactile sensor contains an $8\times14$ pressure grid, yielding 224 values for 2F trials and 336 values for 3F trials. One IMU is mounted on each finger to record accelerometer, gyroscope, and orientation signals. The three snapshots are grouped into one trial-level sample. A trial is labeled as successful if the object remains held when the manipulator reaches the predefined terminal lifting pose at $T_c$, and as failed otherwise. Before quality control, the success/failure counts are 311/189 for 2F and 291/209 for 3F.

\textbf{Data Quality Control:}
All trials are checked for modality completeness, file readability, expected dimensions, numerical validity, and visual-frame quality. A trial is excluded if any required modality at $T_a$, $T_b$, or $T_c$ is missing, unreadable, dimensionally invalid, contains NaN or Inf values, or includes severely corrupted or blurred visual observations. Finite extreme sensor responses are retained because they may reflect genuine contact, impact, or slip events. Of the 1,000 collected trials, two are excluded due to NaN tactile readings caused by temporary sensor disconnection; no other trials are removed. This leaves 998 valid trials, including 600 successful and 398 failed grasps. All quality-control decisions are made before data splitting and without using grasp outcome labels or model predictions.

\section{Methodology}

\subsection{Framework Overview and Design Rationale}

In this section, we introduce SomaAlign, a multimodal representation learning framework for heterogeneous perception in soft robotic grasping. SomaAlign maps vision, depth, IMU, tactile pressure, and text observations into a unified semantic space. Rather than treating multimodal alignment as a uniform feature-matching problem, it explicitly addresses two characteristics of soft robotic perception: modality-dependent reliability and cross-modal semantic inconsistency among observations of the same physical interaction.

Conceptually, SomaAlign is organized around two core methodological designs. \textbf{Reliability-Aware Multimodal Alignment} estimates relative modality reliability and propagates it to modality-pair-specific contrastive alignment, allowing alignment strength to vary across heterogeneous modality pairs. \textbf{Batch-Adaptive Conflict-Aware Geometry-Preserving Refinement} estimates sample-level cross-modal inconsistency, aggregates it at the batch level to regulate refinement strength, and constrains excessive distortion of the global representation geometry. A progressive optimization strategy connects these two designs while retaining only the modality-specific feature encoders for inference.

\begin{figure*}[tb]
	\centering
	\includegraphics[width=\textwidth]{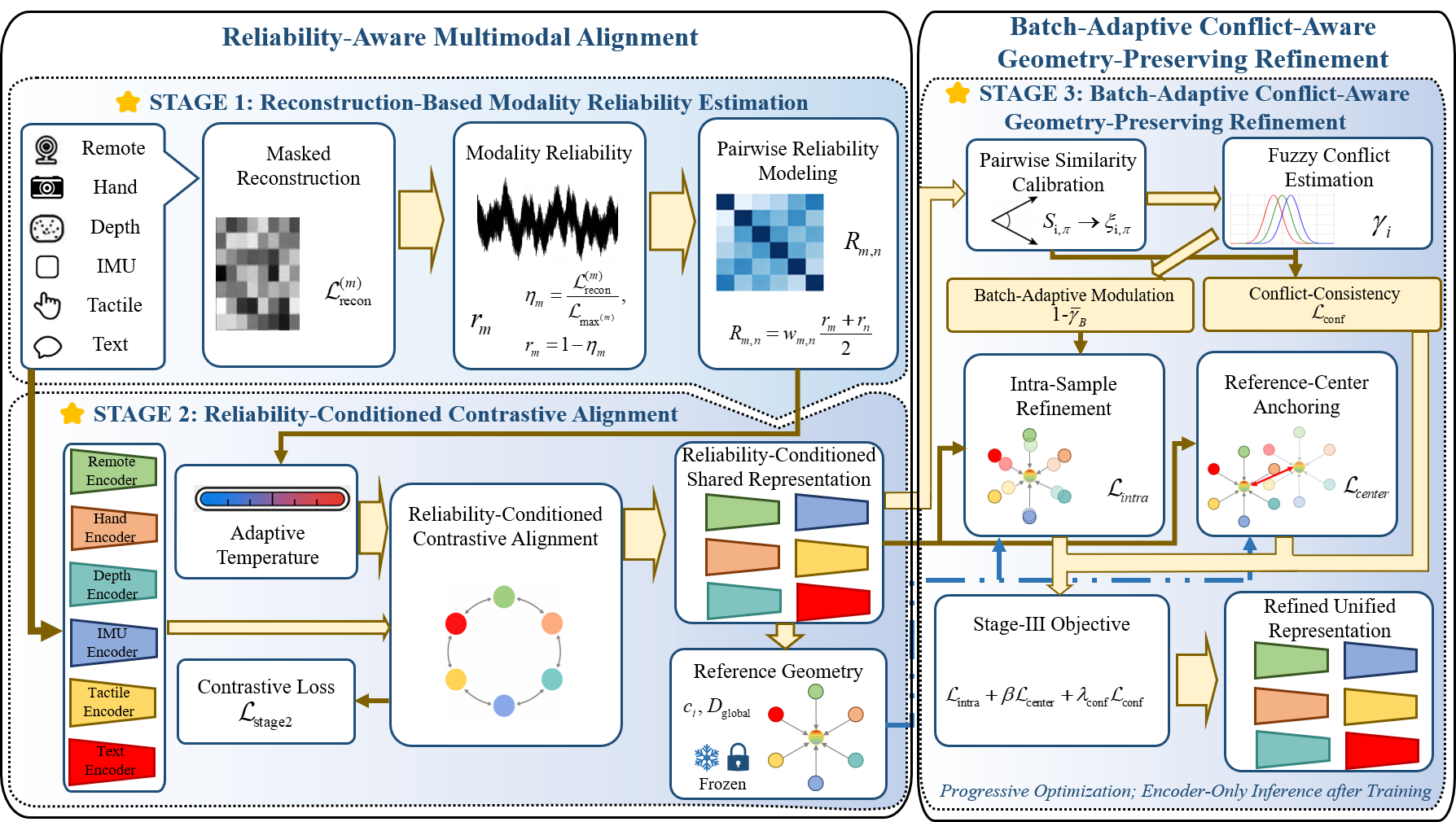}
	\caption{Overview of the SomaAlign framework. SomaAlign follows three progressive optimization stages: (1) Reconstruction-Based Modality Reliability Estimation estimates modality reliability from normalized masked-reconstruction difficulty and constructs pairwise reliability scores; (2) Reliability-Conditioned Contrastive Alignment uses the pairwise reliability scores to adapt cross-modal contrastive temperatures and learn a shared representation; and (3) Batch-Adaptive Conflict-Aware Geometry-Preserving Refinement calibrates pairwise similarities, estimates fuzzy sample-level conflict, and adaptively modulates intra-sample refinement while imposing conflict-consistency and reference-center anchoring constraints. The reference geometry constructed from Stage-II representations, including the sample-specific centers and global representation scale, is frozen and reused during Stage-III optimization. After training, only the modality-specific encoders are retained for inference.}
	\label{Method}
\end{figure*}

The overall workflow is shown in Fig.~\ref{Method}. The two methodological designs are implemented through three progressive optimization stages: Stage I estimates modality reliability and constructs pairwise reliability scores; Stage II performs reliability-conditioned contrastive alignment; and Stage III refines the aligned representation using conflict-aware and geometry-preserving constraints. Thus, the two methodological designs describe the core contributions of SomaAlign, whereas the three stages specify their optimization sequence. After training, all auxiliary components are removed, and raw observations are mapped directly into the unified semantic space by the learned modality-specific encoders.

\subsection{Reliability-Aware Multimodal Alignment}

A fundamental challenge in soft robotic multimodal perception is that different sensing modalities may exhibit substantially different levels of reliability. Recent quality-aware multimodal learning methods have shown that explicitly modeling modality uncertainty or credibility can improve robustness when individual sensory streams are degraded or unreliable \cite{ca14,ca15}. SomaAlign therefore estimates a reconstruction-derived modality reliability proxy before cross-modal alignment and uses it to regulate pair-specific contrastive learning. Tactile pressure and IMU measurements are susceptible to contact deformation, mechanical vibration, and environmental disturbance, whereas visual and textual observations exhibit different and comparatively more structured noise characteristics. Conventional multimodal contrastive learning generally applies the same alignment objective and temperature setting to all modality pairs, which may over-constrain unreliable pairs while underutilizing highly consistent ones.

The reliability-aware alignment design comprises three connected components. In the optimization sequence shown in Fig.~\ref{Method}, reconstruction-based modality reliability estimation and pairwise reliability modeling constitute Stage I, while reliability-conditioned contrastive alignment constitutes Stage II.

\subsubsection{Reconstruction-Based Modality Reliability Estimation}

Because explicit sensor-noise annotations are generally unavailable for heterogeneous multimodal observations, SomaAlign uses masked reconstruction, inspired by masked autoencoding \cite{ca13}, as an auxiliary self-supervised probe of modality-specific reconstruction difficulty. The resulting reliability proxy reflects relative reconstruction stability rather than directly measuring the physical noise level of a sensor: observations with lower predictability or stronger corruption are generally more difficult to recover from partially observed inputs.

For each modality $m$ in a training batch, a random mask is applied to the input $X_m\in\mathbb{R}^{B\times D_m}$, where $B$ is the batch size and $D_m$ is the data dimension. The masked input is

\begin{equation}
	\tilde{X}_m
	=
	X_m\odot M_m,
	\label{e1}
\end{equation}
where $\odot$ denotes the Hadamard product and $M_m\in\{0,1\}^{B\times D_m}$ is a binary mask matrix. Each element of $M_m$ is set to 0 with probability $p_m$ and retained with probability $1-p_m$. The masking probability $p_m$ is selected according to modality characteristics, with higher values for spatially or temporally redundant modalities such as tactile maps and IMU sequences and lower values for information-dense modalities such as text.

A modality-specific reconstruction network $G_m$ reconstructs the original input from $\tilde{X}_m$:

\begin{equation}
	\hat{X}_m
	=
	G_m(\tilde{X}_m),
	\label{eq:reconstruction_output}
\end{equation}
where $\hat{X}_m$ is the reconstructed observation of modality $m$. The architecture of $G_m$ is adapted to each modality: a lightweight convolutional autoencoder is used for visual data, a 2D CNN for tactile pressure maps, a 1D CNN-LSTM for IMU sequences, and a Transformer encoder for text.

Since $M_m=0$ denotes a masked position, we define the complementary mask as $\bar{M}_m=\mathbf{1}-M_m$, where $\mathbf{1}$ denotes an all-ones matrix with the same dimensions as $M_m$. The reconstruction loss is computed only over masked positions:

\begin{equation}
	\mathcal{L}_{\mathrm{recon}}^{(m)}
	=
	\frac{
		\left\|
		\left(\hat{X}_m-X_m\right)
		\odot
		\bar{M}_m
		\right\|_F^2
	}{
		\left\|\bar{M}_m\right\|_1
	},
	\label{eq:recon_loss}
\end{equation}
where $\|\cdot\|_F$ denotes the Frobenius norm and $\|\bar{M}_m\|_1$ denotes the number of masked elements. To make reconstruction difficulty comparable across heterogeneous modalities, a historical maximum reconstruction loss is maintained for each modality. At optimization step $t$,

\begin{equation}
	\mathcal{L}_{\max,t}^{(m)}
	=
	\max\left(
	\mathcal{L}_{\max,t-1}^{(m)},
	\mathcal{L}_{\mathrm{recon},t}^{(m)}
	\right),
	\label{ea1}
\end{equation}
with $\mathcal{L}_{\max,1}^{(m)}=\mathcal{L}_{\mathrm{recon},1}^{(m)}$. Omitting the optimization-step index $t$ hereafter, the normalized reconstruction difficulty is

\begin{equation}
	\eta_m
	=
	\frac{
		\mathcal{L}_{\mathrm{recon}}^{(m)}
	}{
		\mathcal{L}_{\max}^{(m)}
	},
	\qquad
	\eta_m\in[0,1].
	\label{eq:recon_difficulty}
\end{equation}

The corresponding modality reliability is

\begin{equation}
	r_m
	=
	1-\eta_m,
	\label{e4}
\end{equation}
where a larger $r_m$ indicates relatively more stable reconstruction behavior.

\subsubsection{Pairwise Reliability Modeling}

For each modality pair $(m,n)$, the individual modality reliabilities $r_m$ and $r_n$ are combined into a composite reliability score:

\begin{equation}
	R_{m,n}
	=
	w_{m,n}
	\frac{r_m+r_n}{2},
	\label{e5}
\end{equation}
where $w_{m,n}\in(0,1]$ is a fixed prior representing the coarse physical or semantic association strength between modalities $m$ and $n$. We set $w_{m,n}=0.8$ for vision--text pairs and $w_{m,n}=0.5$ for all remaining modality pairs. Importantly, $w_{m,n}$ does not encode sample-specific reliability: the dynamic component of $R_{m,n}$ is determined by the reconstruction-derived reliabilities $r_m$ and $r_n$, while $w_{m,n}$ only modulates the global association strength of the modality pair. The resulting $R_{m,n}$ connects modality-level reliability estimation to pairwise feature alignment.

\subsubsection{Reliability-Conditioned Contrastive Alignment}

Stage II in Fig.~\ref{Method} uses the pairwise reliability scores obtained above to regulate cross-modal contrastive alignment. The temperature parameter in contrastive learning affects the hardness-aware weighting of negative samples and the geometry of the learned representation space \cite{ca21}, while adaptive temperature strategies can condition this parameter on the current alignment or confidence state \cite{ca20}. We therefore use $R_{m,n}$ to generate pair-specific temperatures, allowing contrastive sharpness to vary across heterogeneous modality pairs.

We adopt the UniTouch architecture \cite{c23} with the official pretrained weights as the unified backbone for modality-specific feature extraction. For sample $i$ and modality $m$, the encoder $E_m$ produces

\begin{equation}
	\mathbf{f}_{i,m}
	=
	E_m(X_{i,m})
	\in
	\mathbb{R}^{d},
	\label{eq:encoder_feature}
\end{equation}
where $X_{i,m}$ denotes the observation of modality $m$ for sample $i$, and $d$ is the shared embedding dimension. The feature is $\ell_2$-normalized as

\begin{equation}
	\hat{\mathbf{f}}_{i,m}
	=
	\frac{
		\mathbf{f}_{i,m}
	}{
		\left\|\mathbf{f}_{i,m}\right\|_2+\epsilon
	},
	\qquad
	\epsilon=10^{-8},
	\label{e6}
\end{equation}
where $\|\cdot\|_2$ denotes the Euclidean norm and $\epsilon$ is a small numerical constant used to avoid division by zero.

For each modality pair $(m,n)$, the temperature is determined by $R_{m,n}$:

\begin{equation}
	\tau_{m,n}
	=
	\tau_{\mathrm{base}}
	\exp\left(-\kappa R_{m,n}\right),
	\label{e7}
\end{equation}
where $\tau_{\mathrm{base}}$ is the baseline temperature and $\kappa$ is a learnable scaling coefficient. This monotonic mapping assigns smaller temperatures to more reliable modality pairs, producing sharper similarity distributions, while less reliable pairs receive larger temperatures and smoother distributions. To ensure $\kappa>0$, it is reparameterized as

\begin{equation}
	\kappa
	=
	\exp(\hat{\kappa}),
	\label{eq:kappa_reparameterization}
\end{equation}
where $\hat{\kappa}$ is optimized by backpropagation. A quadratic regularization term $\mathcal{L}_{\mathrm{reg}}$ is applied to $\hat{\kappa}$ to prevent excessive temperature scaling.

Feature alignment is optimized using the InfoNCE objective \cite{ca18}. Averaging over all unique modality pairs gives

\begin{equation}
	\mathcal{L}_{\mathrm{stage2}}
	=
	\frac{2}{M(M-1)}
	\sum_{m=1}^{M-1}
	\sum_{n=m+1}^{M}
	\mathcal{L}_{m,n}
	+
	\lambda_{\mathrm{reg}}
	\mathcal{L}_{\mathrm{reg}},
	\label{e9}
\end{equation}
where $M$ is the number of modalities, $\mathcal{L}_{\mathrm{reg}}$ denotes the quadratic regularization term applied to $\hat{\kappa}$, and $\lambda_{\mathrm{reg}}$ is its weighting coefficient.

For modality pair $(m,n)$,

\begin{equation}
	\mathcal{L}_{m,n}
	=
	-\frac{1}{B}
	\sum_{i=1}^{B}
	\log
	\frac{
		\exp\left(
		\operatorname{sim}
		(\hat{\mathbf{f}}_{i,m},
		\hat{\mathbf{f}}_{i,n})
		/\tau_{m,n}
		\right)
	}{
		\sum_{j=1}^{B}
		\exp\left(
		\operatorname{sim}
		(\hat{\mathbf{f}}_{i,m},
		\hat{\mathbf{f}}_{j,n})
		/\tau_{m,n}
		\right)
	},
	\label{eq:pairwise_infonce}
\end{equation}
where $i$ and $j$ index samples within the same mini-batch. The pair $(\hat{\mathbf{f}}_{i,m},\hat{\mathbf{f}}_{i,n})$ is the positive cross-modal pair, whereas $(\hat{\mathbf{f}}_{i,m},\hat{\mathbf{f}}_{j,n})$ with $j\neq i$ is a negative pair. Because the features are $\ell_2$-normalized, their similarity is

\begin{equation}
	\operatorname{sim}
	(\hat{\mathbf{f}}_{i,m},
	\hat{\mathbf{f}}_{j,n})
	=
	\hat{\mathbf{f}}_{i,m}^{\top}
	\hat{\mathbf{f}}_{j,n}.
	\label{eq:normalized_similarity}
\end{equation}

Thus, the reconstruction-derived pairwise reliability directly controls contrastive sharpness through $R_{m,n}\rightarrow\tau_{m,n}$.

\subsection{Batch-Adaptive Conflict-Aware Geometry-Preserving Refinement}

Although reliability-conditioned contrastive learning establishes a globally aligned multimodal representation space, cross-modal consistency may still vary across samples because of transient contact conditions, motion disturbances, partial visual observations, and semantic ambiguity. SomaAlign therefore introduces batch-adaptive conflict-aware geometry-preserving refinement, corresponding to Stage III in Fig.~\ref{Method}. It combines sample-level conflict estimation, batch-level adaptive modulation, and frozen Stage-II reference geometry to improve local cross-modal consistency without excessively distorting the global representation structure.

\subsubsection{Reference Geometry Construction}

After Stage II, let $\mathbf{f}_{i,m}\in\mathbb{R}^{d}$ denote the aligned feature vector of sample $i$ from modality $m$. A frozen reference center is constructed for each training sample:

\begin{equation}
	\mathbf{c}_i
	=
	\frac{1}{M}
	\sum_{m=1}^{M}
	\mathbf{f}_{i,m},
	\label{e11}
\end{equation}
where $\mathbf{c}_i\in\mathbb{R}^{d}$ is the Stage-II reference center of sample $i$. These centers characterize the sample-level geometry established by Stage-II alignment and remain fixed throughout Stage III.

To normalize center displacement relative to the overall representation scale, we compute the mean pairwise distance between all reference centers:

\begin{equation}
	D_{\mathrm{global}}
	=
	\frac{2}{N(N-1)}
	\sum_{i<j}
	\left\|
	\mathbf{c}_i-\mathbf{c}_j
	\right\|_2,
	\label{e12}
\end{equation}
where $N$ is the number of training samples used to construct the reference geometry. Both $\{\mathbf{c}_i\}_{i=1}^{N}$ and the scalar $D_{\mathrm{global}}$ are computed once from the Stage-II representations and kept fixed during Stage III.

\subsubsection{Distribution-Calibrated Fuzzy Conflict Estimation}

Because heterogeneous modality pairs may exhibit substantially different similarity distributions, directly comparing their raw similarities can introduce pair-dependent scale bias. We therefore calibrate each modality pair against its own Stage-II similarity distribution before estimating cross-modal conflict.

Let $\pi$ index a modality pair $(m,n)$, with $\pi\in\{1,\ldots,P\}$ and $P=M(M-1)/2$ denoting the total number of unique modality pairs. For sample $i$, the cosine similarity associated with pair $\pi$ is

\begin{equation}
	s_{i,\pi}
	=
	\frac{
		\mathbf{f}_{i,m}^{\top}\mathbf{f}_{i,n}
	}{
		\|\mathbf{f}_{i,m}\|_2
		\|\mathbf{f}_{i,n}\|_2
	},
	\label{e13}
\end{equation}
where $\mathbf{f}_{i,m}$ and $\mathbf{f}_{i,n}$ denote the current feature vectors of the two modalities associated with pair $\pi$. They are initialized from the Stage-II aligned representations and progressively updated during Stage-III optimization. 

For each modality pair $\pi$, an empirical cumulative distribution function (ECDF) $F_{\pi}(\cdot)$ is estimated exclusively from the fixed Stage-II representations of the training split. The raw similarity is converted into a calibrated pairwise conflict indicator:

\begin{equation}
	\xi_{i,\pi}
	=
	1-F_{\pi}(s_{i,\pi}),
	\label{ea2}
\end{equation}
where a larger $\xi_{i,\pi}$ indicates stronger inconsistency relative to the typical similarity distribution of modality pair $\pi$. All pair-specific ECDFs remain fixed throughout Stage III.

The calibrated indicator is processed by $K_{\mathrm{f}}$ ordered Gaussian membership functions:

\begin{equation}
	\mu_{i,k,\pi}
	=
	\exp
	\left[
	-\frac{
		(\xi_{i,\pi}-a_k)^2
	}{
		2\sigma_k^2
	}
	\right],
	\label{eq:fuzzy_membership}
\end{equation}
where $k\in\{1,\ldots,K_{\mathrm{f}}\}$ indexes the fuzzy rules, and $a_k$ and $\sigma_k$ denote the center and width of the $k$-th Gaussian membership function, respectively.

\begin{figure}[tb]
	\centering
	\includegraphics[width=\columnwidth]{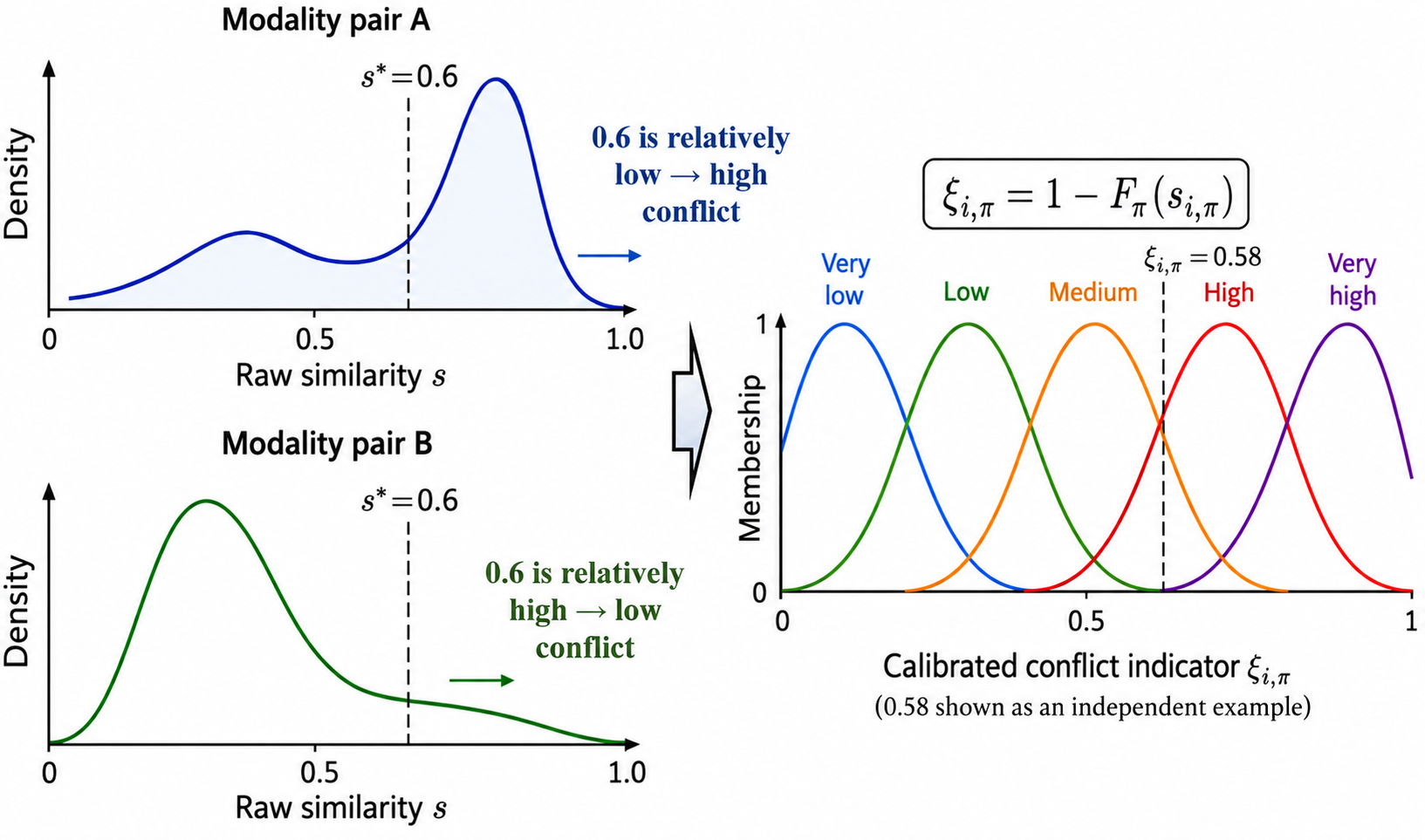}
	\caption{Conceptual illustration of pair-specific distribution calibration and fuzzy membership assignment. Different modality pairs may exhibit different raw similarity distributions; therefore, the same similarity value can represent different degrees of relative inconsistency. Pair-specific ECDFs convert raw similarities into comparable conflict indicators $\xi_{i,\pi}$. The calibrated indicator is then softly assigned to overlapping ordered Gaussian fuzzy sets, allowing transitional cross-modal conflicts to partially activate multiple adjacent conflict levels.}
	\label{fig:fuzzy_conflict}
\end{figure}

As illustrated in Fig.~\ref{fig:fuzzy_conflict}, the memberships are normalized across the rule dimension to obtain fuzzy-rule weights:

\begin{equation}
	\omega_{i,k,\pi}
	=
	\frac{
		\mu_{i,k,\pi}
	}{
		\sum_{\ell=1}^{K_{\mathrm{f}}}
		\mu_{i,\ell,\pi}
	},
	\label{eq:fuzzy_rule_weight}
\end{equation}
where $\ell$ is a summation index over the fuzzy rules and $\sum_{k=1}^{K_{\mathrm{f}}}\omega_{i,k,\pi}=1$. The normalized rule weights are combined with ordered consequents $v_k$ to obtain a continuous pair-level conflict score:

\begin{equation}
	q_{i,\pi}
	=
	\sum_{k=1}^{K_{\mathrm{f}}}
	\omega_{i,k,\pi}v_k,
	\label{eq:pair_conflict_score}
\end{equation}
where $v_k$ is the consequent of the $k$-th fuzzy rule and increases from low to high conflict. Consequently, stronger activation of high-conflict rules produces a larger $q_{i,\pi}$.

To capture both overall inconsistency and a small number of strongly conflicting modality pairs, the pair-level scores are aggregated as

\begin{equation}
	q_i
	=
	(1-\rho)
	\operatorname{Mean}_{\pi}(q_{i,\pi})
	+
	\rho
	\operatorname{TopKMean}_{\pi}(q_{i,\pi};K_{\mathrm{t}}),
	\label{e16}
\end{equation}
where $\rho\in[0,1]$ controls the emphasis on salient pairwise conflicts. The global mean is

\begin{equation}
	\operatorname{Mean}_{\pi}(q_{i,\pi})
	=
	\frac{1}{P}
	\sum_{\pi=1}^{P}
	q_{i,\pi},
	\label{eq:pairwise_mean}
\end{equation}
and $\operatorname{TopKMean}_{\pi}(q_{i,\pi};K_{\mathrm{t}})$ denotes the average of the $K_{\mathrm{t}}$ largest values among $\{q_{i,\pi}\}_{\pi=1}^{P}$. Thus, $\rho=0$ uses only the global mean, whereas $\rho=1$ uses only the $K_{\mathrm{t}}$ most salient modality-pair conflicts.

The aggregated score is finally mapped to a bounded sample-level conflict estimate:

\begin{equation}
	\gamma_i
	=
	\operatorname{sigmoid}
	\left[
	\alpha(q_i-b)
	\right],
	\label{e17}
\end{equation}
where $\alpha$ and $b$ are the scale and offset of the sigmoid mapping, respectively. Rather than setting them manually, we calibrate them using lower and upper quantiles $Q_{\mathrm{l}}$ and $Q_{\mathrm{u}}$ of the training-set distribution of $q_i$:

\begin{equation}
	\alpha
	=
	\frac{
		\operatorname{logit}(\gamma_{\mathrm{u}})
		-
		\operatorname{logit}(\gamma_{\mathrm{l}})
	}{
		Q_{\mathrm{u}}-Q_{\mathrm{l}}
	},
	\qquad
	b
	=
	Q_{\mathrm{l}}
	-
	\frac{
		\operatorname{logit}(\gamma_{\mathrm{l}})
	}{
		\alpha
	},
	\label{ea4}
\end{equation}
where $\gamma_{\mathrm{l}}$ and $\gamma_{\mathrm{u}}$ are the prescribed target conflict levels corresponding to $Q_{\mathrm{l}}$ and $Q_{\mathrm{u}}$, respectively, and $\operatorname{logit}(z)=\log\frac{z}{1-z}$ is the inverse of the sigmoid function.

\subsubsection{Batch-Adaptive Conflict Modulation}

The sample-level conflict estimates $\{\gamma_i\}$ are aggregated to obtain a batch-level conflict factor:

\begin{equation}
	\bar{\gamma}_{B}
	=
	\frac{1}{B}
	\sum_{i=1}^{B}
	\operatorname{sg}(\gamma_i),
	\label{ea7}
\end{equation}
where $\operatorname{sg}(\cdot)$ denotes the stop-gradient operator. The scalar $\bar{\gamma}_{B}$ summarizes the overall cross-modal conflict level of the current batch and modulates the strength of representation refinement.

Let $\mathbf{f}_{i,m}^{\mathrm{III}}\in\mathbb{R}^{d}$ denote the current Stage-III feature vector of sample $i$ from modality $m$. The conflict-modulated intra-sample refinement objective is

\begin{equation}
	\mathcal{L}_{\mathrm{intra}}
	=
	(1-\bar{\gamma}_{B})
	\frac{1}{B}
	\sum_{i=1}^{B}
	\frac{1}{M}
	\sum_{m=1}^{M}
	\left\|
	\mathbf{f}_{i,m}^{\mathrm{III}}
	-
	\mathbf{c}_i
	\right\|_2,
	\label{e18}
\end{equation}
where $\mathbf{c}_i$ is the frozen Stage-II reference center defined above. A lower batch-level conflict strengthens the contraction toward the Stage-II reference centers, whereas a higher conflict weakens it. Because all samples in the batch share the same modulation factor $1-\bar{\gamma}_{B}$, their relative weighting within $\mathcal{L}_{\mathrm{intra}}$ is preserved.

\subsubsection{Conflict-Consistency Learning}

To provide an explicit learning signal for the fuzzy conflict scorer, we construct a calibrated target from the ECDF-normalized pairwise conflict indicators:

\begin{equation}
	t_i
	=
	(1-\rho)
	\operatorname{Mean}_{\pi}(\xi_{i,\pi})
	+
	\rho
	\operatorname{TopKMean}_{\pi}
	(\xi_{i,\pi};K_{\mathrm{t}}),
	\label{ea5}
\end{equation}
where $t_i$ is the conflict target for sample $i$, using the same aggregation operators, $\rho$, and $K_{\mathrm{t}}$ defined above. The fuzzy conflict scorer is optimized by matching the predicted sample-level conflict $\gamma_i$ to this stop-gradient target:

\begin{equation}
	\mathcal{L}_{\mathrm{conf}}
	=
	\frac{1}{B}
	\sum_{i=1}^{B}
	\operatorname{SmoothL1}
	\left(
	\gamma_i,
	\operatorname{sg}(t_i)
	\right),
	\label{ea6}
\end{equation}
where $\operatorname{SmoothL1}(\cdot,\cdot)$ denotes the standard smooth-$L_1$ (Huber-type) regression loss. Thus, $\mathcal{L}_{\mathrm{conf}}$ trains the fuzzy scorer to reproduce the calibrated conflict structure without propagating gradients through the target construction.

\subsubsection{Reference-Center Anchoring for Geometry Preservation}

Although intra-sample refinement promotes local cross-modal compactness, excessive displacement of sample centers may distort the global geometry established in Stage II. We therefore define the current Stage-III center of sample $i$ as

\begin{equation}
	\bar{\mathbf{c}}_i
	=
	\frac{1}{M}
	\sum_{m=1}^{M}
	\mathbf{f}_{i,m}^{\mathrm{III}},
	\label{eq:stage3_center}
\end{equation}
where $\bar{\mathbf{c}}_i\in\mathbb{R}^{d}$ denotes the current Stage-III center relative to the frozen Stage-II reference center $\mathbf{c}_i$.

\begin{figure}[tb]
	\centering
	\includegraphics[width=\columnwidth]{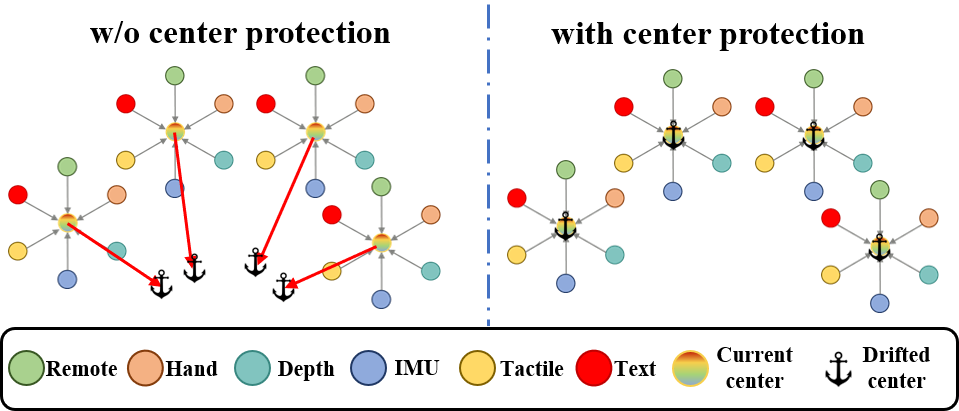}
	\caption{Motivation of the center-protection loss. Local cross-modal refinement may cause sample centers to deviate from the geometry established in Stage II. The center-protection constraint penalizes excessive center drift while allowing local multimodal representations to become more compact.}
	\label{fig:center_protection_motivation}
\end{figure}

As illustrated in Fig.~\ref{fig:center_protection_motivation}, the center-protection loss is activated only when the displacement from the frozen reference exceeds a scale-adaptive tolerance:

\begin{equation}
	\mathcal{L}_{\mathrm{center}}
	=
	\frac{1}{B}
	\sum_{i=1}^{B}
	\max
	\left(
	0,
	\left\|
	\bar{\mathbf{c}}_i-\mathbf{c}_i
	\right\|_2
	-
	\lambda_{\mathrm{tol}}D_{\mathrm{global}}
	\right),
	\label{e19}
\end{equation}
where $\lambda_{\mathrm{tol}}$ controls the allowable center displacement relative to the global representation scale $D_{\mathrm{global}}$. It determines the activation margin of the center-protection penalty rather than its loss weight.

The geometry-preserving effect can be further characterized by the triangle inequality. Let

\begin{equation}
	d_{ij}
	=
	\left\|
	\mathbf{c}_i-\mathbf{c}_j
	\right\|_2
	\label{eq:reference_center_distance}
\end{equation}
denote the Stage-II distance between the frozen reference centers of samples $i$ and $j$. If the refined centers satisfy

\begin{equation}
	\begin{aligned}
		\left\|
		\bar{\mathbf{c}}_i-\mathbf{c}_i
		\right\|_2
		&\leq
		\lambda_{\mathrm{tol}}D_{\mathrm{global}},\\
		\left\|
		\bar{\mathbf{c}}_j-\mathbf{c}_j
		\right\|_2
		&\leq
		\lambda_{\mathrm{tol}}D_{\mathrm{global}},
	\end{aligned}
	\label{eq:center_tolerance_condition}
\end{equation}
then

\begin{equation}
	\left\|
	\bar{\mathbf{c}}_i-\bar{\mathbf{c}}_j
	\right\|_2
	\geq
	d_{ij}
	-
	2\lambda_{\mathrm{tol}}D_{\mathrm{global}}.
	\label{e20}
\end{equation}
Thus, bounding each Stage-III center around its frozen Stage-II reference limits excessive distortion of inter-sample geometry while still permitting local multimodal refinement.

\subsubsection{Overall Conflict-Aware Refinement Objective}

The complete objective used for Stage-III refinement is

\begin{equation}
	\mathcal{L}_{\mathrm{stage3}}
	=
	\mathcal{L}_{\mathrm{intra}}
	+
	\beta\mathcal{L}_{\mathrm{center}}
	+
	\lambda_{\mathrm{conf}}
	\mathcal{L}_{\mathrm{conf}},
	\label{e21}
\end{equation}
where $\beta$ is the weight of the geometry-preservation loss and $\lambda_{\mathrm{conf}}$ is the weight of the conflict-consistency loss. Unlike $\beta$, which controls the magnitude of the center-protection penalty, $\lambda_{\mathrm{tol}}$ defines the center-drift margin at which $\mathcal{L}_{\mathrm{center}}$ becomes active.

\subsection{Progressive Optimization and Encoder-Only Inference}

The two methodological designs above are connected through the progressive optimization strategy summarized in Algorithm~\ref{alg:somaalign}.

\begin{algorithm}[tb]
	\caption{SomaAlign Progressive Multimodal Alignment}
	\label{alg:somaalign}
	\begin{algorithmic}[1]
		\renewcommand{\algorithmicrequire}{\textbf{Input:}}
		\renewcommand{\algorithmicensure}{\textbf{Output:}}
		\REQUIRE Dataset $\mathcal{D}$; number of modalities $M$; stage epoch counts $N_{\mathrm{ep}}^{(1)}$, $N_{\mathrm{ep}}^{(2)}$, and $N_{\mathrm{ep}}^{(3)}$.
		\ENSURE Fine-tuned modality encoders $\{E_m\}_{m=1}^{M}$.
		
		\STATE \textbf{Stage I: Reliability Estimation}
		\FOR{$epoch=1$ to $N_{\mathrm{ep}}^{(1)}$}
		\STATE Optimize each reconstruction network $G_m$ using $\mathcal{L}_{\mathrm{recon}}^{(m)}$
		\STATE Update modality-specific historical reconstruction references
		\ENDFOR
		\STATE Estimate $\{r_m\}$ and construct pairwise reliabilities $\{R_{m,n}\}$
		
		\STATE \textbf{Stage II: Reliability-Conditioned Alignment}
		\FOR{$epoch=1$ to $N_{\mathrm{ep}}^{(2)}$}
		\STATE Compute pair-specific temperatures $\{\tau_{m,n}\}$
		\STATE Optimize modality encoders using $\mathcal{L}_{\mathrm{stage2}}$
		\ENDFOR
		
		\STATE Construct and freeze $\{\mathbf{c}_i\}$, $D_{\mathrm{global}}$, and pair-specific ECDFs
		\STATE Calibrate and freeze sigmoid parameters $\alpha$ and $b$
		
		\STATE \textbf{Stage III: Batch-Adaptive Conflict Refinement}
		\FOR{$epoch=1$ to $N_{\mathrm{ep}}^{(3)}$}
		\STATE Estimate sample conflicts $\{\gamma_i\}$ by calibrated fuzzy inference
		\STATE Compute batch conflict factor $\bar{\gamma}_{B}$
		\STATE Compute $\mathcal{L}_{\mathrm{intra}}$, $\mathcal{L}_{\mathrm{conf}}$, and $\mathcal{L}_{\mathrm{center}}$
		\STATE Update modality encoders and fuzzy conflict scorer using $\mathcal{L}_{\mathrm{stage3}}$
		\ENDFOR
		
		\STATE \textbf{return} $\{E_m\}_{m=1}^{M}$
	\end{algorithmic}
\end{algorithm}

Quantities used as statistical or geometric references in Stage III---including the Stage-II reference centers, $D_{\mathrm{global}}$, pair-specific ECDFs, and sigmoid calibration parameters---are constructed from the training split after Stage II and remain fixed throughout Stage-III optimization. The modality encoders and fuzzy conflict scorer are optimized using these fixed references.

All auxiliary mechanisms are training-only. After optimization, the reconstruction networks, reliability estimation and temperature modulation, ECDF calibration, fuzzy conflict scorer, batch-level conflict modulation, and reference-center constraints are discarded. Only the fine-tuned modality-specific encoders $\{E_m\}_{m=1}^{M}$ are retained for inference.

\section{Experiments}

\subsection{Experimental Setup}

SomaAlign is implemented in PyTorch with modality-specific encoders initialized from UniTouch pretrained weights. Experiments are conducted on NVIDIA RTX 4090 GPUs, each modified to provide 48\,GB of VRAM, using AdamW and a cosine annealing learning-rate schedule. Stage~I is trained for 50 epochs with a batch size of 64. Stages~II and~III are each trained for 200 epochs with an initial learning rate of $1\times10^{-5}$, a per-GPU batch size of 8, and four gradient-accumulation steps.

For Stage~I, the masking probabilities $p_m$ are fixed at 0.7 for tactile and IMU, 0.6 for both RGB streams and depth, and 0.3 for text. For Stage~II, we set $\tau_{\mathrm{base}}=0.3$ and $\lambda_{\mathrm{reg}}=0.1$. The temperature scaling coefficient $\kappa$ is initialized to 1 by setting $\hat{\kappa}=0$, and $\hat{\kappa}$ is optimized jointly with the encoders.

For Stage~III, we use $K_{\mathrm{f}}=5$ ordered Gaussian fuzzy rules, with centers $a_k$ initialized to $\{0.1,0.3,0.5,0.7,0.9\}$, widths $\sigma_k$ initialized to $0.15$, and consequents $v_k$ initialized to $\{0,0.25,0.5,0.75,1\}$. Conflict aggregation in Eq.~\eqref{e16} uses $\rho=0.5$ and $K_{\mathrm{t}}=3$. The 5th and 95th percentiles of the training-set distribution of $q_i$ are mapped to target conflict levels $\gamma_{\mathrm{l}}=0.2$ and $\gamma_{\mathrm{u}}=0.8$, respectively. Calibration using Eq.~\eqref{ea4} yields $\alpha=5.10$ and $b=0.582$. The center-drift tolerance and loss weights are $\lambda_{\mathrm{tol}}=0.1$, $\beta=0.25$, and $\lambda_{\mathrm{conf}}=0.1$. The reference centers, global distance scale, pair-specific ECDFs, and sigmoid calibration parameters are constructed using only the Stage-II training representations and remain fixed throughout Stage~III.

Experiments on Soma follow the 60\%/20\%/20\% training/validation/test split defined in Section~II-C, with each grasping trajectory assigned entirely to one subset. The training subset is used for representation learning and reference construction, the validation subset for model and hyperparameter selection, and the test subset for final evaluation. The independent Soma-Success evaluation protocol is specified in its corresponding subsection. Additional parameter-effect and diagnostic analyses are provided in the Supplementary Information.

\subsection{Cross-Modal Retrieval Performance}

Cross-modal retrieval evaluates correspondence in the shared representation space by ranking gallery features according to their cosine similarity to a query. For query $q$, let $\mathcal{P}_q$ denote its valid positive set. Recall@$K$ is defined as
\begin{equation}
	\mathrm{Recall@}K = \frac{1}{Q}\sum_{q=1}^{Q}\mathbb{I}\left(\operatorname{TopK}(q)\cap\mathcal{P}_q\neq\emptyset\right),
	\label{e22}
\end{equation}
where $Q$ is the number of queries, $\operatorname{TopK}(q)$ contains the top-$K$ retrieved gallery entries, and $\mathbb{I}(\cdot)$ is the indicator function.

For sensory-to-sensory retrieval, $\mathcal{P}_q$ contains only the paired observation from the same temporal window. For sensory-to-text retrieval, the gallery contains one entry per unique structured description, and the query's corresponding description is the positive match. For text-to-sensory retrieval, each unique description is queried once, and all sensory observations sharing exactly that description are valid positives. Thus, sensory-only retrieval uses observation-level correspondence, whereas text-related retrieval uses description-level correspondence.

We evaluate all 30 directed retrieval tasks among the six modalities on a fixed 500-sample subset of the test split. Recall@$K$ is computed separately for each direction and then macro-averaged, giving equal weight to all directions despite differences in query and gallery sizes. SomaAlign achieves mean Recall@1, Recall@5, and Recall@10 values of 63.04\%, 85.55\%, and 90.66\%, respectively.

\begin{figure*}[tb]
	\centering
	\includegraphics[width=\linewidth]{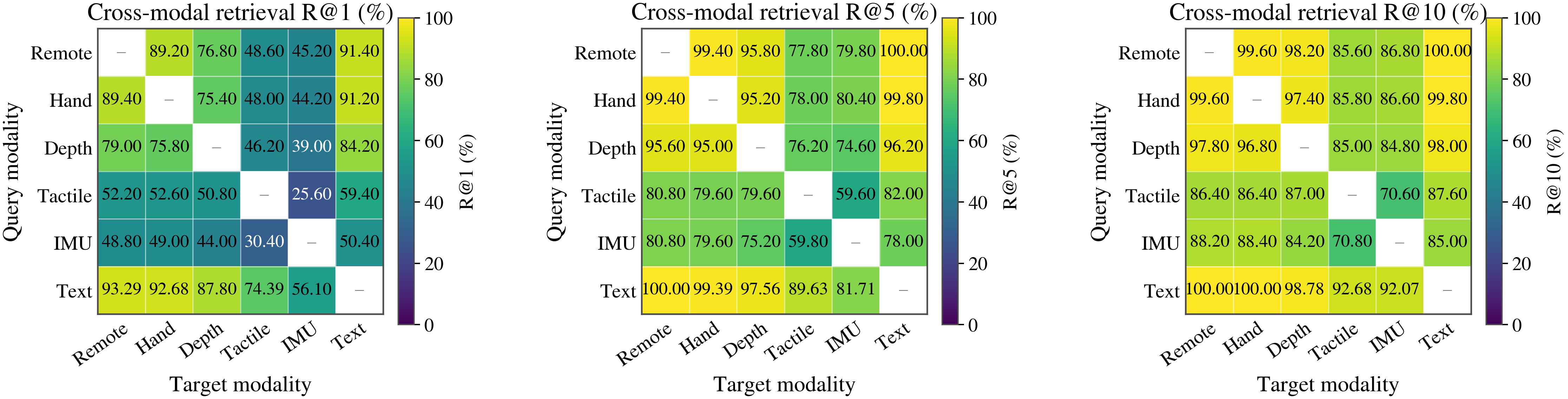}
	\caption{Recall@1, Recall@5, and Recall@10 across 30 directed cross-modal retrieval tasks. Sensory-only retrieval uses paired temporal-window correspondence; text-related retrieval uses identical structured descriptions. Remote and Hand denote remote-view and hand-mounted RGB, respectively. Same-modality retrieval is excluded.}
	\label{fig:retrieval_heatmap}
\end{figure*}

As shown in Fig.~\ref{fig:retrieval_heatmap}, Remote$\rightarrow$Hand and Remote$\rightarrow$Text achieve Recall@10 values of 99.60\% and 100.00\%, respectively. Contact-related modalities are more challenging at the top rank: Tactile$\rightarrow$Text and IMU$\rightarrow$Text achieve Recall@1 values of 59.40\% and 50.40\%, increasing to 87.60\% and 85.00\% at Recall@10. These results indicate that the aligned representations recover cross-modal correspondence, while top-ranked matching remains less reliable for tactile and IMU observations than for visual observations.

\textbf{Comparison with Controlled Alignment Baselines.}
We compare SomaAlign with two controlled alignment baselines: Pairwise InfoNCE~\cite{ca18}, which aligns all modality pairs, and ImageBind-style Hub Alignment~\cite{ca7}, which aligns each of the other five modalities with Remote RGB as the visual hub. Both controlled baselines use the same modality encoders and UniTouch pretrained initialization as SomaAlign. Their encoder-training configurations, including the optimizer, learning rate and schedule, batch configuration, and training budget, are matched to those of SomaAlign. All three methods follow the same data split and evaluation protocol. Both baselines use a fixed temperature of $\tau=0.15$ without reconstruction-based reliability estimation, reliability-conditioned temperature modulation, or Stage-III refinement.

\begin{table}[tb]
	\centering
	\caption{Comparison with controlled multimodal alignment baselines on Soma. All Recall values are percentages.}
	\label{tab:controlled_baselines}
	\begin{tabular}{cccc}
		\hline
		Method & R@1 & R@5 & R@10 \\
		\hline
		Pairwise InfoNCE~\cite{ca18} & 42.00 & 60.10 & 65.77 \\
		Hub Alignment~\cite{ca7} & 44.55 & 64.30 & 70.55 \\
		\textbf{SomaAlign} & \textbf{63.04} & \textbf{85.55} & \textbf{90.66} \\
		\hline
	\end{tabular}
\end{table}

As shown in Table~\ref{tab:controlled_baselines}, SomaAlign outperforms both baselines across all three retrieval metrics, improving R@1 by 21.04 and 18.49 percentage points over Pairwise InfoNCE and Hub Alignment, respectively.

To assess training variability, we evaluate three independently trained Pairwise InfoNCE checkpoints on the same fixed test subset under the retrieval protocol defined above. The mean $\pm$ sample standard deviation is 43.72$\pm$1.62\%, 63.38$\pm$3.09\%, and 69.53$\pm$3.54\% for R@1, R@5, and R@10, respectively. Separately, evaluating one fixed SomaAlign checkpoint on five independently sampled 500-instance test subsets yields 60.78$\pm$1.64\%, 84.03$\pm$1.13\%, and 89.11$\pm$1.11\%. The primary results in Table~\ref{tab:controlled_baselines} and Fig.~\ref{fig:retrieval_heatmap} are reported on the same fixed 500-instance test subset.

\textbf{Comparison with Published Multimodal Methods.}
We compare SomaAlign with TVL~\cite{c15}, TLV-Link~\cite{ca19}, and UniTouch~\cite{c23}, adapted to a common vision--tactile--language (V--T--L) retrieval setting on Soma using Hand RGB, tactile pressure, and structured English text. All methods follow the same experiment-level data split and are evaluated across the six directed V--T--L retrieval tasks using the correspondence rules defined above. SomaAlign retains six-modality training and is evaluated here on the shared V--T--L interface.

\begin{table}[tb]
	\centering
	\caption{Comparison with published multimodal methods under the common V--T--L retrieval protocol on Soma. Recall values are percentages macro-averaged over six directed tasks.}
	\label{tab:published_vtl}
	\begin{tabular}{cccc}
		\hline
		Method & Avg. R@1 & Avg. R@5 & Avg. R@10 \\
		\hline
		TVL-Adapted~\cite{c15} & 55.64 & 75.09 & 81.02 \\
		TLV-Link-Adapted~\cite{ca19} & 40.51 & 60.65 & 69.89 \\
		UniTouch-Adapted~\cite{c23} & 48.16 & 64.47 & 76.86 \\
		\textbf{SomaAlign} & \textbf{69.71} & \textbf{88.07} & \textbf{92.05} \\
		\hline
	\end{tabular}
\end{table}

As shown in Table~\ref{tab:published_vtl}, SomaAlign achieves the highest average Recall across all three metrics, exceeding TVL-Adapted, the strongest baseline by average retrieval performance, by 14.07 and 11.03 percentage points in R@1 and R@10, respectively.

\textbf{Robustness to Sensor Corruption.}
We evaluate retrieval robustness by adding Gaussian noise to tactile or IMU observations:
\begin{equation}
	\tilde{\mathbf{x}}=\mathbf{x}+\nu\sigma(\mathbf{x})\mathbf{z},\qquad
	\mathbf{z}\sim\mathcal{N}(\mathbf{0},\mathbf{I}),
	\label{eq:sensor_corruption}
\end{equation}
where $\sigma(\mathbf{x})$ is the standard deviation of the clean observation and $\nu$ controls the noise magnitude relative to this scale. We test $\nu\in\{0,0.01,0.03,0.05,0.1,0.3,0.7\}$, perturbing one modality at a time while keeping the others unchanged. SomaAlign, Pairwise InfoNCE, and Hub Alignment are evaluated under identical corruption settings, with results averaged over five noise seeds using fixed trained checkpoints and test samples.

Across the tested corruption levels, SomaAlign maintains higher R@1 and R@10 than both controlled baselines. At $\nu=0.7$, SomaAlign retains 75.1\%/73.4\% of its clean R@1/R@10 under tactile corruption, compared with 70.3\%/70.0\% for Pairwise InfoNCE and 71.0\%/70.7\% for Hub Alignment. Under IMU corruption, the corresponding retention rates are 78.1\%/72.9\%, 73.8\%/68.5\%, and 73.2\%/69.1\%, respectively. Detailed results across all corruption levels are provided in the Supplementary Information.

\subsection{Downstream Classification Performance}

We evaluate frozen 1024-dimensional non-text features extracted by SomaAlign using logistic regression and an MLP with a 512--256 architecture and ReLU activations. Structured annotations provide the target labels, while text features are excluded from the probe inputs. Independent classifiers are trained for hardness, grasping state, material, size, and shape, with data split by experiment ID to keep grasping trajectories disjoint across subsets.

Across the five tasks, the linear probe achieves a mean accuracy of 93.84\% and a mean F1-score of 92.62\%, while the MLP reaches 95.45\% accuracy and 94.03\% F1-score. Detailed per-task results are provided in the Supplementary Information.

We further compare Pairwise InfoNCE and a Uniform Prior variant under the same frozen-feature probing protocol. Uniform Prior sets $w_{m,n}=1$ for all modality pairs while retaining the reconstruction-derived modality reliabilities and the remaining training pipeline.

\begin{table}[tb]
	\centering
	\caption{Mean probe accuracy across the five classification tasks (\%).}
	\label{tab:probe_variant_comparison}
	\setlength{\tabcolsep}{4pt}
	\begin{tabular}{ccc}
		\toprule
		\textbf{Representation} & \textbf{Linear} & \textbf{MLP} \\
		\midrule
		Pairwise InfoNCE & 92.06 & 94.40 \\
		Uniform Prior & 92.89 & 94.90 \\
		\textbf{SomaAlign} & \textbf{93.84} & \textbf{95.45} \\
		\bottomrule
	\end{tabular}
\end{table}

As shown in Table~\ref{tab:probe_variant_comparison}, SomaAlign achieves the highest mean accuracy with both probe architectures, exceeding Pairwise InfoNCE and Uniform Prior by 1.78 and 0.95 percentage points with the linear probe and by 1.05 and 0.55 percentage points with the MLP, respectively.

\subsection{Frozen-Encoder Evaluation on Soma-Success}

We evaluate the reuse of Soma-trained representations for grasp-outcome prediction on the independently collected Soma-Success dataset, which is not used during representation learning. All modality encoders remain frozen, and only linear or MLP probes are trained on Soma-Success.

For each trial, $\mathbf{x}_{t,m}$ denotes the observation of modality $m$ at stage $t\in\{\mathrm{pre},\mathrm{mid},\mathrm{post}\}$, corresponding to $T_a$, $T_b$, and $T_c$, respectively. The same stage-selection rule is applied independently of success/failure labels. Features from the available modalities are aggregated as
\begin{equation}
	\mathbf{z}_{t}=\operatorname{Agg}\left(\{E_m(\mathbf{x}_{t,m})\mid m\in\mathcal{M}_{t}\}\right),
	\label{eq:success_stage_feature}
\end{equation}
where $\mathcal{M}_{t}$ is the set of available modalities at stage $t$ and $\operatorname{Agg}(\cdot)$ denotes the feature aggregation operation used by the probe.

Post-hoc classification uses all three observation stages:
\begin{equation}
	\mathbf{z}_{\mathrm{all}}=[\mathbf{z}_{\mathrm{pre}};\mathbf{z}_{\mathrm{mid}};\mathbf{z}_{\mathrm{post}}],
	\label{eq:success_posthoc_feature}
\end{equation}
whereas early prediction uses only pre-grasp and mid-grasp observations:
\begin{equation}
	\mathbf{z}_{\mathrm{early}}=[\mathbf{z}_{\mathrm{pre}};\mathbf{z}_{\mathrm{mid}}],
	\label{eq:success_early_feature}
\end{equation}
where $[\cdot;\cdot]$ denotes concatenation. A probe with parameters $\phi$ predicts the binary outcome:
\begin{equation}
	\hat{y}=h_{\phi}(\mathbf{z}_{s}),\qquad s\in\{\mathrm{all},\mathrm{early}\},
	\label{eq:success_probe_prediction}
\end{equation}
with labels $y=0$ for failure and $y=1$ for success.
%

Linear and two-layer MLP probes are evaluated using five-fold cross-validation grouped by object identity. All trials associated with the same object, including both gripper configurations, are assigned to the same fold. In each iteration, the probes are trained on four folds and evaluated on the remaining fold, while all modality encoders remain frozen. Each trial therefore receives one out-of-fold prediction from a probe trained without any trials of that object. After quality control, the dataset contains 998 valid trials, comprising 600 successful and 398 failed grasps.

\begin{table}[tb]
	\centering
	\caption{Frozen-encoder grasp-outcome classification accuracy on Soma-Success (\%), reported as mean $\pm$ standard deviation over five folds.}
	\label{tab:soma_success_comparison}
	\setlength{\tabcolsep}{2.2pt}
	\begin{tabular}{ccccc}
		\toprule
		\textbf{Representation} & \multicolumn{2}{c}{\textbf{Post-hoc}} & \multicolumn{2}{c}{\textbf{Early}} \\
		\cmidrule(lr){2-3} \cmidrule(lr){4-5}
		& \textbf{Linear} & \textbf{MLP} & \textbf{Linear} & \textbf{MLP} \\
		\midrule
		Pairwise InfoNCE & 89.78$\pm$1.95 & 90.28$\pm$1.43 & 82.87$\pm$2.61 & 84.07$\pm$3.24 \\
		Uniform Prior & 92.49$\pm$1.64 & 92.38$\pm$1.16 & 82.87$\pm$1.75 & 85.07$\pm$1.84 \\
		Hub Alignment & 89.18$\pm$1.06 & 90.98$\pm$0.70 & 81.67$\pm$2.19 & \textbf{86.48$\pm$2.52} \\
		\textbf{SomaAlign} & \textbf{93.18$\pm$2.18} & \textbf{94.59$\pm$1.94} & \textbf{83.07$\pm$2.60} & 84.07$\pm$1.55 \\
		\bottomrule
	\end{tabular}
\end{table}

%

As shown in Table~\ref{tab:soma_success_comparison}, SomaAlign achieves the highest mean post-hoc accuracy with both probes: 93.18\% for the linear probe and 94.59\% for the MLP. These results support the reuse of its frozen representations for grasp-outcome classification using all three observation stages.

Early-prediction performance remains competitive when only pre-grasp and mid-grasp observations are available. SomaAlign achieves the highest mean linear accuracy of 83.07\%, while its MLP probe reaches 84.07\%, compared with 86.48\% for the best-performing Hub Alignment. These results indicate that the frozen SomaAlign representations retain useful predictive information even before the post-grasp observation becomes available. Confusion matrices obtained by pooling the out-of-fold predictions across all five folds are provided in the Supplementary Information.

\subsection{Cross-Modal Generation Analysis}

We conduct an auxiliary leave-one-out generation experiment to assess whether aligned features from the remaining modalities support reconstruction of a missing stream. For each target modality, a conditional diffusion model uses features from the other five modalities as conditioning inputs. For visual, depth, and tactile modalities, generation is performed in a VAE-compressed latent space, whereas IMU sequences are modeled directly with a 1D U-Net due to their low-dimensional temporal structure. Qualitative generation examples are provided in the Supplementary Information.

We further evaluate attribute-level semantic preservation using two linear-probe strategies. \textbf{Single-Modality Probe} trains and tests the classifier using only features from the generated target modality. \textbf{Multi-Modality Probe} trains the classifier on averaged features from all six real modalities and replaces only the target modality with its generated counterpart at test time.

\begin{figure*}[tb]
	\centering
	\includegraphics[width=\linewidth]{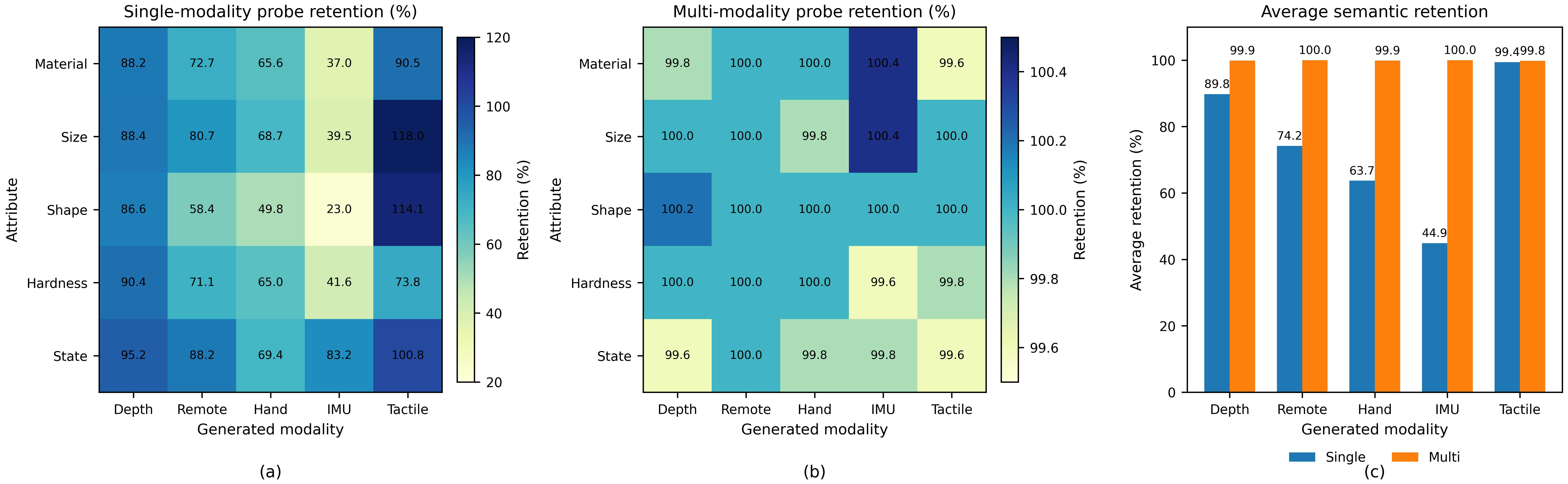}
	\caption{Attribute classification accuracy retention under cross-modal generation, measured relative to the corresponding real-data baseline. (a) Single-modality probes. (b) Multi-modality probes with one generated modality replacing its real counterpart. (c) Average retention for the two strategies.}
	\label{fig:generation_retention}
\end{figure*}

Accuracy retention is defined as classification accuracy under the generated-data condition divided by that under the corresponding real-data condition. As shown in Fig.~\ref{fig:generation_retention}, single-modality retention varies across targets: Depth achieves the highest retention, the two RGB streams show intermediate retention, and IMU achieves the lowest at 44.9\%. Tactile also exhibits high retention, although its relatively low real-data baseline accuracy limits the absolute classification performance implied by this ratio.

Average multi-modality retention remains at 99.8\%--100.0\% across the evaluated target modalities. Thus, replacing one modality with its generated counterpart leaves fused classification performance close to the all-real baseline, despite larger losses in standalone semantic retention for some modalities.

\subsection{Multimodal Question Answering}

We further evaluate whether SomaAlign representations support language-based reasoning over soft robotic multimodal observations. A LLaMA-based language model~\cite{ca9} is fine-tuned using the aligned features from Remote RGB, Hand RGB, Depth, Tactile, and IMU observations. QA-aware augmentation constructs 44 training variants per sample, including one original structured description and 43 question/instruction--answer variants covering object descriptions, identity, individual attributes, grasping state, and multi-attribute queries.

For quantitative evaluation, the model generates a structured description for each of the 15,085 Soma test samples. Exact Match is computed by case-insensitive full-string comparison with the reference description; normalized Edit Distance and BLEU-4 are additionally reported.

\begin{table}[tb]
	\centering
	\caption{Overall multimodal question-answering metrics on the Soma test split.}
	\label{tab:gen_overall}
	\begin{tabular}{lc}
		\toprule
		\textbf{Metric} & \textbf{Value} \\
		\midrule
		Test samples & 15,085 \\
		Exact Match & 92.0\% \\
		Mean normalized Edit Distance & 0.0059 \\
		Mean BLEU-4 & 0.9878 \\
		\bottomrule
	\end{tabular}
\end{table}

As shown in Table~\ref{tab:gen_overall}, 13,880 of the 15,085 generated descriptions exactly match their references. The mean normalized Edit Distance of 0.0059 and BLEU-4 score of 0.9878 further indicate high agreement with the reference descriptions. A complementary field-level evaluation yields a pooled attribute accuracy of 98.41\%. Detailed field-level results, gripper-specific statistics, and qualitative question-answering examples are provided in the Supplementary Information.

\subsection{Ablation Studies}

We evaluate five core variants: (1) w/o Stage I removes reconstruction-based reliability estimation and uses uniform pairwise reliability; (2) w/o Adaptive Temp replaces reliability-conditioned temperature modulation with $\tau=0.15$; (3) w/o Stage~III removes the entire conflict-aware refinement stage; (4) w/o Center loss removes $\mathcal{L}_{\mathrm{center}}$; and (5) w/o Intra loss removes $\mathcal{L}_{\mathrm{intra}}$. Other hyperparameters are kept unchanged except where required by the ablation.

\begin{table*}[tb]
	\centering
	\caption{Ablation of the core components in SomaAlign.}
	\label{tab:ablation_components}
	\begin{tabular}{lcccccc}
		\toprule
		\textbf{Model} & \textbf{R@1 (\%)} & \textbf{R@5 (\%)} & \textbf{R@10 (\%)} & \textbf{k-NN (\%)} & \textbf{Fisher} & \textbf{CHI} \\
		\midrule
		SomaAlign & \textbf{63.04} & \textbf{85.55} & \textbf{90.66} & \textbf{87.2} & 1967.1 & 7.318 \\
		\midrule
		w/o Stage I & 47.94 & 73.92 & 85.79 & 85.6 & 1863.9 & 7.241 \\
		w/o Adaptive Temp ($\tau=0.15$) & 47.27 & 73.63 & 84.67 & 86.6 & 1937.9 & 7.168 \\
		w/o Stage III & 22.72 & 50.85 & 68.45 & 86.4 & 2220.7 & 7.315 \\
		w/o Center loss & 34.48 & 65.26 & 73.64 & 86.2 & 1928.0 & 6.282 \\
		w/o Intra loss & 26.98 & 57.51 & 65.17 & 86.2 & \textbf{2261.7} & \textbf{7.661} \\
		\bottomrule
	\end{tabular}
\end{table*}

Table~\ref{tab:ablation_components} shows that removing Stage~I or replacing adaptive temperature modulation reduces R@1 by 15.10 and 15.77 percentage points, respectively. Both variants also reduce R@5 and R@10, supporting the contribution of reconstruction-derived reliability and its use in temperature modulation.

Removing Stage~III produces the largest R@1 decrease among the core variants, at 40.32 percentage points. Removing its center-protection or intra-sample refinement loss decreases R@1 by 28.56 and 36.06 percentage points, respectively. These results support the contribution of both geometric refinement losses. Notably, w/o Stage~III and w/o Intra loss achieve higher Fisher scores than SomaAlign, and the latter also achieves a higher CHI. Their lower retrieval performance shows that these separability metrics alone do not capture cross-modal correspondence.

\textbf{Effect of the Modality-Pair Prior.}
The Uniform Prior variant defined above achieves R@1, R@5, and R@10 values of 45.47\%, 67.41\%, and 74.98\%, respectively, with $w_{m,n}=1$ for all modality pairs. Its R@1 is 17.57 percentage points below SomaAlign, supporting the use of the specified modality-pair priors in the reliability-conditioned alignment scheme.

\textbf{Effect of Conflict Modulation Strategy.}
We compare four strategies while keeping the remaining Stage-III components unchanged. No-Conflict uses an intra-sample refinement coefficient of 1. Fixed-Weight uses a constant matched to the average coefficient of Mean-Gamma. Sample-Gamma applies the sample-specific coefficient $1-\operatorname{sg}(\gamma_i)$, whereas Mean-Gamma uses the batch-level coefficient $1-\bar{\gamma}_{B}$ adopted in SomaAlign.

\begin{table}[tb]
	\centering
	\caption{Ablation of Stage-III conflict modulation strategies. Recall values are percentages.}
	\label{tab:conflict_modulation}
	\begin{tabular}{cccc}
		\hline
		Method & R@1 & R@5 & R@10 \\
		\hline
		No-Conflict & 54.44 & 80.72 & 86.38 \\
		Fixed-Weight & 58.61 & 83.57 & 89.39 \\
		Sample-Gamma & 59.00 & 83.48 & 89.05 \\
		\textbf{Mean-Gamma} & \textbf{63.04} & \textbf{85.55} & \textbf{90.66} \\
		\hline
	\end{tabular}
\end{table}

Mean-Gamma achieves the highest Recall across all three metrics (Table~\ref{tab:conflict_modulation}), improving R@1 by 8.60, 4.43, and 4.04 percentage points over No-Conflict, Fixed-Weight, and Sample-Gamma, respectively. Fixed-Weight and Sample-Gamma perform similarly. These results favor batch-level adaptation of refinement strength, which applies a shared coefficient while preserving relative sample weighting within the intra-sample loss.

%
%
%

\subsection{Inference Complexity and Runtime}

SomaAlign representation inference retains only the modality-specific encoders required for the six sensing modalities, resulting in 1,332.2M active inference parameters. Encoding one six-modality sample requires 1.666 TFLOPs, and FP16 inference takes 43.67 ms per sample on the evaluated hardware. Detailed runtime, batch-processing, and parameter-storage statistics are provided in the Supplementary Information.

\subsection{Discussion}

The experimental findings suggest that multimodal representation learning for soft robotic perception requires both accommodating modality-dependent reliability and preserving meaningful cross-modal correspondence during feature alignment. Soft robotic sensing is affected by deformation-dependent contact responses, transient mechanical disturbances, and heterogeneous modality-specific noise, making it difficult to assume that all modality pairs should be aligned with identical strength. The results therefore support a progressive strategy in which modality reliability is first estimated, then incorporated into cross-modal optimization, and finally complemented by geometry-preserving semantic refinement.

An important implication of the reliability-aware design is that modality reliability need not be treated only as a feature-level weighting factor. In SomaAlign, reconstruction-derived reliability modulates the temperature of pairwise contrastive learning and therefore changes the sharpness of the alignment objective itself. More reliable modality pairs can receive stronger alignment constraints, whereas less reliable pairs are optimized more conservatively. This is particularly relevant for heterogeneous visual, tactile, inertial, and semantic observations whose stability and noise characteristics can differ substantially. Reliability thus serves not only as an estimate of modality stability, but also as a mechanism for regulating how strongly different modalities are brought into a shared representation space.

The experiments also highlight that global feature separability and cross-modal correspondence are distinct representation properties. A feature space may exhibit favorable class-level organization while still failing to correctly associate heterogeneous observations originating from the same physical interaction. This distinction is important for multimodal robotic perception, where semantic organization and observation-level correspondence are both necessary. The Stage-III design can be interpreted from this perspective: intra-sample refinement promotes local cross-modal compactness, whereas center protection limits excessive displacement from the Stage-II reference geometry. These objectives therefore address complementary aspects of representation refinement, balancing local consistency against preservation of the broader feature structure.

The conflict-modulation results provide a further methodological observation. Sample-level conflict estimates are useful indicators of the current multimodal consistency state, but they need not directly define relative sample importance. Directly applying sample-specific conflict weights may couple optimization too strongly to instantaneous local inconsistency. By contrast, batch-level aggregation allows the overall refinement strength to adapt to the current cross-modal state while preserving the relative contributions of individual samples within the batch. This suggests that conflict information is more appropriately used as an optimization-state signal than as a direct sample-weighting criterion.

The downstream evaluations further indicate that multimodal representation quality is inherently multidimensional. Cross-modal retrieval measures correspondence across modalities, physical-attribute probes assess semantic accessibility, Soma-Success evaluates the reusability of frozen representations on independently collected data, and generation and question answering provide additional views of the information retained in the shared representation. Differences across these tasks therefore reflect distinct representation properties rather than a single notion of performance. In particular, the Soma-Success results show that the learned features remain useful under both complete and partial observation settings, although downstream performance can vary with the available observations and probe architecture. This also suggests that representation quality should be assessed through complementary tasks rather than inferred from any single metric.

Several aspects of the current framework define the scope of these conclusions. The Stage-I reliability score is derived from masked reconstruction and should therefore be interpreted as a reconstruction-based proxy for relative modality stability rather than a direct measurement of physical sensor noise. The uniform-masking diagnostic suggests that the observed cross-modal differences are not determined solely by predefined masking probabilities, while explicit sensor calibration or uncertainty modeling could provide more physically grounded estimates. The structured text annotations provide consistent semantic supervision but limit linguistic diversity, motivating future extensions toward more open-ended descriptions and question formulations. In addition, Soma-Success demonstrates frozen-encoder reuse with lightweight supervised probes rather than zero-shot generalization, and broader evaluation across platforms, sensor configurations, and end-effectors would further characterize transferability. Extending SomaAlign toward action-conditioned prediction, robot world models~\cite{ca10,ca11}, and closed-loop soft robotic control is a natural next step.

\section{Conclusion}

We presented Soma, Soma-Success, and SomaAlign for multimodal perception in soft robotic grasping. Soma provides software-coordinated, cycle-aligned Remote RGB, Hand RGB, depth, tactile, and IMU observations together with structured text descriptions. SomaAlign combines reliability-aware multimodal alignment with batch-adaptive conflict-aware geometry-preserving refinement to learn a shared representation of these heterogeneous observations.

Experiments demonstrate the utility of the learned representations for cross-modal retrieval and physical attribute recognition, with ablations supporting the contributions of reliability-conditioned alignment and conflict-aware refinement. Additional generation and question-answering experiments illustrate downstream uses of the aligned features. Frozen-encoder evaluation on the independently collected Soma-Success dataset yields the highest mean post-hoc grasp-outcome accuracy among the evaluated representations, while early-prediction performance varies with probe architecture.

Future work will extend the data coverage to more diverse manipulation scenarios, incorporate physically grounded reliability estimates, and explore the use of SomaAlign representations in action-conditioned prediction and closed-loop soft robotic control.

\vfill

\end{document}